\documentclass[runningheads]{llncs}

\PassOptionsToPackage{dvipsnames,table}{xcolor}
 
\usepackage{eccvabbrv}

\usepackage{graphicx}
\usepackage{booktabs}
\usepackage{longtable} 
\usepackage{multirow}
\usepackage{tcolorbox}
\tcbuselibrary{breakable}
\usepackage{fvextra}
\usepackage{algorithm}
\usepackage{algorithmic}
\usepackage{enumitem}
\newcommand{\RETURN}{\STATE \textbf{return} }

\usepackage[accsupp]{axessibility}  

\usepackage[nolist]{acronym}

\usepackage{hyperref}

\usepackage{orcidlink}

\begin{document}

\begin{acronym}[LVQA]
\acro{VLM}{vision-language model}
\acro{LLM}{large language model}
\acro{RL}{reinforcement learning}
\acro{GRPO}{Group Relative Policy Optimization}
\acro{PPO}{Proximal Policy Optimization}
\acro{GSPO}{Group Sequence Policy Optimization}
\acro{RLOO}{REINFORCE Leave-One-Out}
\acro{SFT}{supervised fine-tuning}
\acro{RM}{reward model}
\acro{RAG}{retrieval-augmented generation}
\acro{TL}{temporal logic}
\acro{LTL}{linear temporal logic}
\acro{MDP}{Markov decision process}
\acro{LVQA}{long-video question answering}
\acro{VQA}{visual question answering}
\acro{QA}{question answering}
\acro{ID}{in-domain}
\acro{OOD}{out-of-domain}
\acro{EM}{exact match}
\acro{VETL}[\texttt{VETL}]{Visual Evidence via Temporal Logic}
\end{acronym}

\title{Incentivizing Vision Language Models to Search for Long Video Question Answering} 

\titlerunning{VSeek}

\newcommand{\samethanks}[1][\value{footnote}]{\footnotemark[#1]}
\author{Harsh Goel\thanks{Equal Contribution.}\orcidlink{0009-0006-9873-9584}\and
S P Sharan\samethanks\orcidlink{0000-0002-6298-6464}\and
Sahil Shah\orcidlink{0009-0009-3640-2339}\and
Minkyu Choi\orcidlink{0009-0007-6557-8865}\and
Joungbin An\orcidlink{0000-0003-0418-900X}\and
Kristen Grauman\orcidlink{0000-0002-9591-5873}\and
Sandeep P. Chinchali\orcidlink{0000-0002-0601-3633}
}

\authorrunning{H.~Goel et al.}

\institute{The University of Texas at Austin, Austin, TX, USA\\
}

\maketitle

\begin{abstract}
    We introduce \texttt{VSeek}, an agentic framework that transforms \ac{LVQA} from a passive, single-pass perception task into a multi-turn retrieval process. \texttt{VSeek} utilizes a natural language-driven search to identify relevant context within long videos and is post-trained with \ac{RL} to jointly formulate targeted search queries and reason over retrieved clips for \ac{LVQA}. While \ac{RL} post-training has revolutionized reasoning in symbolic domains such as mathematics and code, its application to long-video understanding remains hindered by a lack of verified rewards. To ensure that the retrieved context is relevant, we propose a novel neuro-symbolic approach that bridges open-ended natural language with discrete visual verification. Specifically, complex user queries are compiled into formal temporal logic specifications for systematically decomposing natural language questions into a definitive checklist of required atomic visual primitives, such as key objects and activities, along with their temporal ordering. These systematically derived grounding events provide the critical feedback signal for \ac{RL} post-training, enabling dense, verifiable rewards based on the successful retrieval of these specific visual elements rather than relying entirely on outcome-only answer accuracy. By explicitly optimizing for this verifiable evidence-seeking behavior, \texttt{VSeek} improves Pass@1 scores by up to 8\% and Pass@4 scores by 15\% on long-video understanding benchmarks compared to base models. We open-source our code at \url{https://utaustin-swarmlab.github.io/VSeek}.
    
\keywords{Video Question and Answering \and Multimodal Reasoning \and Vision Language Models \and Reinforcement Learning}
\end{abstract}

\section{Introduction}

\begin{figure}[t]
        \centering
        \includegraphics[width=0.9\linewidth]{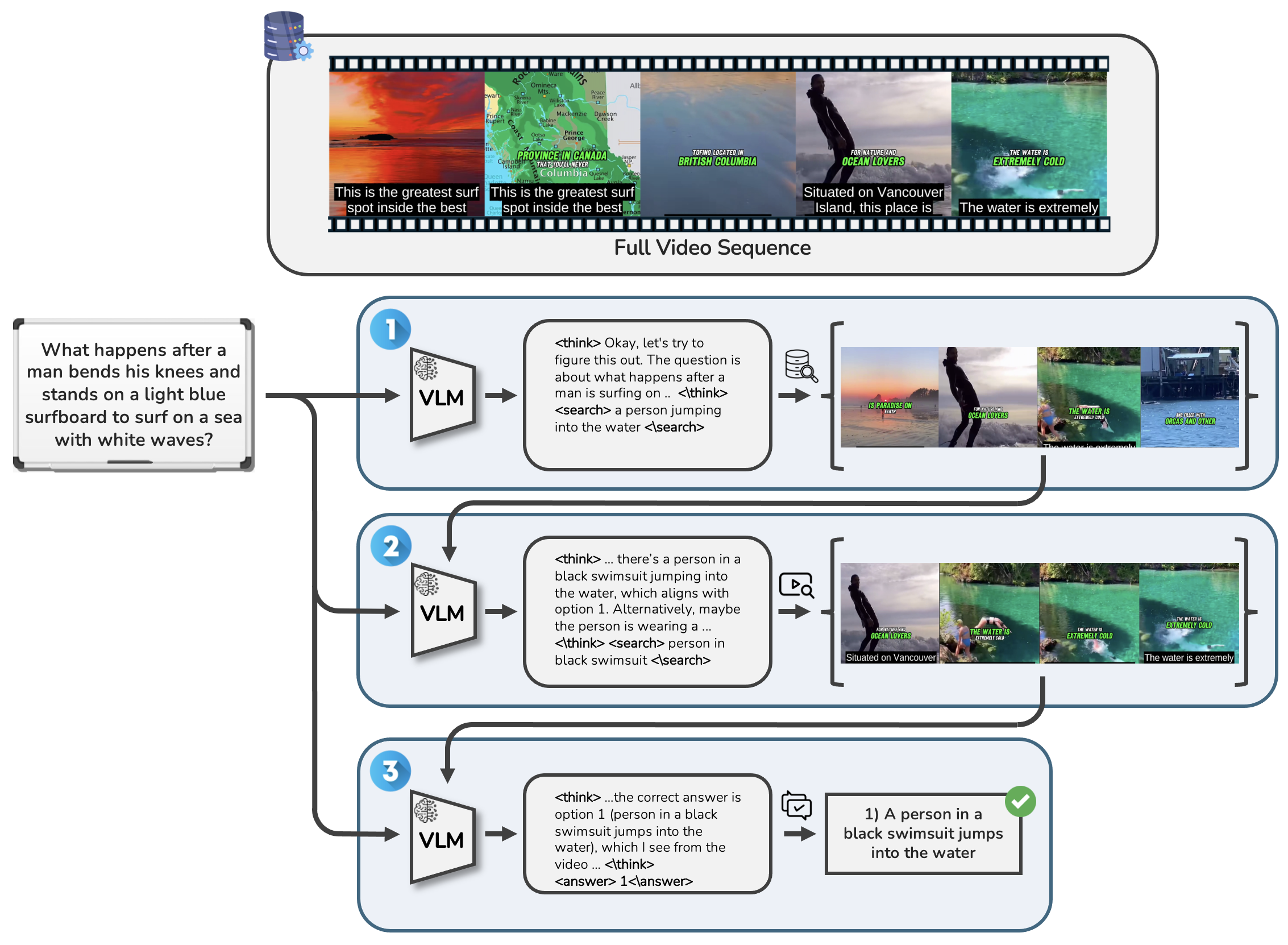}
        \caption{\textbf{Visualization of VSeek}. Given an indexed long video and a question, \texttt{VSeek} iteratively decomposes the query, issues natural-language search requests to retrieve candidate clips, and refines its video context before producing the final answer.}
        \label{fig:video agent and puls}
\end{figure}

When humans are tasked with answering questions pertaining to a two-hour documentary, they do not rely on a sparse, random sampling of frames; instead, they scrub through the timeline to identify relevant segments. Despite this, current post-training paradigms for video-language models \cite{bai2025qwen25vltechnicalreport,li2025videochat,feng2025videor1,bai2025qwen3} fundamentally train models over uniformly sampled frames from long videos. This is insufficient for long-video understanding since uniform sampling inevitably misses the sparse, query-critical moments \cite{shah2025neus,wang2025videotree,ye2025rethinking} required to answer complex queries while utilizing irrelevant context that leads to incorrect answers \cite{hong2025context}.

To tackle this challenge, recent work has explored acquiring relevant video context by equipping models with tools to retrieve specific clips \cite{wang2025videotree,wang2024videoagent,jeoung2024adaptive,tan2025allvballinonelongvideo,islam2025bimba,shah2025neus}, which entirely rely on prompting off-the-shelf foundation models to identify relevant frames from per-frame video captions. Other works have explicitly trained dedicated tool-calling vision-language models \cite{he2025framethinker,yuan2025videoexplorer} to teach the model to identify and output precise time intervals to obtain relevant video clips through extensive \ac{SFT} before \ac{RL} post-training. To bypass the limitations of exhaustive caption scanning and rigid temporal prediction, we propose \texttt{VSeek}, an agentic framework that actively searches for query-specific context with natural language as a primary search interface (see Figure \ref{fig:video agent and puls}). However, current pipelines rely almost exclusively on ``outcome-only'' supervision \cite{bai2025qwen25vltechnicalreport,bai2025qwen3} alongside auxiliary objectives like temporal ordering \cite{feng2025videor1,xue2026seeing}, which would fail to certify whether an agent has constructed a query-specific context when using natural language-based video search.


We argue that the path forward for agentic long-video \ac{QA} must mirror the recent breakthroughs in \ac{LLM} reasoning for mathematics \cite{shao2025deepseekmath,shao2024deepseekmath} and programming \cite{roziere2023code,guo2024deepseek,team2024codegemma}, where progress is catalyzed by verifiable rewards (\textit{i.e.}, unit tests and formal symbolic rules). However, applying this level of exact rule-checking to raw video is difficult because natural language queries inherently entangle the semantic content of visual events with complex temporal dependencies. This is necessary to score the query-specific contexts built by video-understanding agents. Thus, in this work, we also introduce \ac{VETL}, a pipeline to decouple visual evidence from its temporal dependencies in natural language questions, to formulate a reward signal. By translating these open-ended queries into formal temporal logic specifications, we systematically disentangle the underlying visual grounding events or primitives and represent their temporal relationships through temporal operators.
Thus, long-horizon questions are transformed into an exact checklist of visual prerequisites (\textit{i.e.}, entities, actions, and relations) that act as visual ``unit tests'' defining valid query-specific contexts. Through \ac{VETL}, we obtain rewards for \ac{RL} post-training for open-ended visual context construction, therefore improving the agentic search and answering process of \texttt{VSeek}.

Our contributions are summarized as follows:
\begin{itemize}
    \item We introduce \textbf{\texttt{VSeek}}, an agentic framework that turns long-video \ac{QA} into an active, multi-turn discovery process by decomposing complex queries and issuing targeted natural-language searches for relevant clips.
    \item We propose \ac{VETL}, a neuro-symbolic pipeline that closes the verification gap by converting questions into video-grounded primitives and compiling them into verifiable temporal-logic specifications that formalize the ``correct context'' for any query.
    \item We design \ac{RL} rewards from \ac{VETL} that score an agent's evidence-gathering trajectory against the resulting extracted primitives, enabling checkable stepwise feedback beyond outcome-only answer supervision.
    \item We show that optimizing for verified context construction boosts long-horizon agentic reasoning: \texttt{VSeek} improves Pass@1 by up to 8\% and Pass@4 by 15\% over base models on long-video understanding benchmarks.
\end{itemize}

\section{Related Work}

\subsection{Video Question Answering}

Early \ac{VQA} systems~\cite{antol2015vqa,ye2017video} encoded entire videos through convolutional, recurrent or attention-based pipelines~\cite{jang2017tgif,zhu2016visual7w}, but scaled poorly to longer inputs \cite{huang2020movienet,tapaswi2016movieqa,chandrasegaran2025hourvideo}. Modern \acp{VLM}~\cite{achiam2023gpt4,bai2025qwen25vltechnicalreport,li2024llava,opengvlab2024internvl} enable zero-shot \ac{VQA} with stronger generalization, yet these still operate over a fixed set of uniformly sampled frames. This dilutes relevant context in multi-event sequences~\cite{wang2024lvbench,wu2024longvideobench,song2025moviechat+,huang2020movienet} and degrades performance on queries requiring precise temporal localization.

Retrieval-augmented methods~\cite{islam2025bimba,wang2024videoagent,wang2025videotree,ye2025rethinking,he2024malmm} mitigate this by conditioning on task-relevant segments, but typically rely on coarse heuristics or textual summaries that sacrifice visual fidelity. Dedicated tool-calling \acp{VLM}~\cite{he2025framethinker,yuan2025videoexplorer} offer more structured video grounding, yet their rigid interfaces demand extensive \ac{SFT} to learn format-specific commands and timestamp mappings. \texttt{VSeek} sidesteps both limitations by using flexible natural-language queries as the search interface, enabling an intuitive search without format-specific supervision.

\subsection{Agentic Search and RL Post-Training}

Text-based agentic frameworks~\cite{schick2023toolformer,yao2023react,goel2026s,jin2025searchr1} interleave reasoning with tool calls; systems such as Search-R1~\cite{jin2025searchr1} encapsulate thought chains and integrate retrieved evidence for better \ac{RAG}~\cite{lewis2020rag}. These pipelines excel in knowledge-intensive text tasks, but extending them to video, where evidence is visual and not textual, remains unsolved. \texttt{VSeek} addresses this by treating long-video understanding as a temporal search-and-retrieval task.

\Ac{RL} post-training through \ac{GRPO}~\cite{shao2024deepseekmath,guo2025deepseek} has driven rapid gains in math and code, where verifiable rewards (unit tests, compilation checks) provide dense and reliable training signals~\cite{cobbe2021verifiers,guo2024deepseek,team2024codegemma}. Transferring this to video is hard: existing pipelines use outcome-only supervision~\cite{yuan2025videoexplorer,feng2025videor1} or self-verification~\cite{pan2025timesearch}, and therefore do not directly incentivize correct context construction to prevent spurious shortcuts~\cite{amodei2016concrete}. \texttt{VSeek} fills this gap by compiling queries into temporal-logic specifications~\cite{murphy2024guiding,pnueli1977temporal} that act as visual ``unit tests,'' providing rewards that incentivize correct evidence gathering rather than final-answer accuracy alone.

\subsection{Neuro-Symbolic Methods for Video Understanding}

Extracting symbolic representations from video objects, actions, relations, and events has been explored for classification~\cite{feichtenhofer2019slowfast,tran2019channel}, event detection~\cite{li2022anomalous,medioni2001event}, \ac{VQA}~\cite{chen2022comphy}, robotic planning~\cite{hasanbeig2019reinforcement,kress2009temporal}, and autonomous driving verification~\cite{jha2018safe,mehdipour2023formal}. These symbols are typically obtained from spatio-temporal graphs~\cite{sarkar2015early,yu2022probabilistic} or latent-space~\cite{bertasius2021spacetime,kroshchanka2021neural}, and translated into formal languages such as temporal logic~\cite{Baier2008}. Within video understanding, symbolic methods have supported long-form reasoning~\cite{huang2020movienet,tapaswi2016movieqa}, and recent work~\cite{yang2023specification,choi2024towards} represents videos as formal models to search for events. To represent queries for evaluation, grounding, and question-answering, recent work \cite{shah2025neus,munir2025objectalign,sharan2025neuro,choi2025we,liang2026neus} extracts video-based primitives in temporal logic for subsequent neuro-symbolic verification. Crucially, all prior neuro-symbolic methods use temporal logic for \emph{post-hoc} context construction; none use them as a \emph{training signal} \cite{shah2025challenge}. \texttt{VSeek} departs from this by compiling queries into formal specifications over visual primitives and using them to \emph{shape dense rewards} during \ac{RL} post-training, turning open-ended context construction into a verifiable objective.
\section{Preliminaries}

\subsection{Search with LLMs}
\label{sec:prelim_multihop}

We draw inspiration from text-based search agents such as Search-R1 \cite{jin2025searchr1} and R1-Searcher \cite{song2025r1}, which auto-regressively generates a complete response $y \sim \pi_{\theta}(\cdot \mid x)$, given a prompt $x$ and a search engine, as an interleaved sequence of reasoning and tool calls. The policy $\pi_{\theta}$ decodes $y$ autoregressively, so that the sequence-level distribution factorizes into per-token conditionals, $\pi_{\theta}(y \mid x) = \prod_{t} \pi_{\theta}(y_t \mid x, y_{<t})$; we use $\pi_{\theta}$ to denote both forms interchangeably throughout. Guided by a strict token protocol, these models reason within \texttt{<think>} tags, issue queries via \texttt{<search>}, and iteratively assimilate returned evidence or context within \texttt{<information>} tags until they produce a final answer within \texttt{<answer>} tags. We extend this paradigm to the video domain, transforming long-video understanding from a passive, single-pass perception problem into an \textbf{agentic retrieval} task. By explicitly interleaving active search with internal thinking, the model can dynamically construct context to produce cogent answers on longer videos.




\subsection{Group Relative Policy Optimization for Multi-hop Search}
\label{sec:prelim_grpo}
\ac{RL} methods optimize \ac{LLM} policies with task-specific rewards. While \ac{PPO} \cite{schulman2017proximal} uses a learned value function, \ac{GRPO} \cite{shao2024deepseekmath} eliminates its need by standardizing rewards within a sampled group.

The core principle of \ac{GRPO} is to establish a relative baseline for policy updates using the average reward from a group of sampled trajectories. For each input prompt $x$ from the dataset $\mathcal{D}$, \ac{GRPO} samples a group of $G$ responses $\{y_1, y_2, \dots, y_G\}$ from a policy $\pi_{\text{old}}$ to optimize the policy $\pi_{\theta}$ through the objective 
\begin{equation}\label{eq:grpo_final}
J_{\text{GRPO}}(\theta) = \mathbb{E}_{x \sim \mathcal{D}, \{y_i\}_{i=1}^G \sim \pi_{\text{old}}} \Biggl[ \frac{1}{G} \sum_{i=1}^{G} \frac{\sum_{t: I(y_{i,t})=1} \Bigl( \mathcal{L}_{\text{clip}}(\theta,t) - \beta \mathcal{D}_{\text{KL}}(\theta,t) \Bigr)}{\sum_{t} I(y_{i,t})} \Biggr],\end{equation}
where the clipped surrogate per-token loss $\mathcal{L}_{\text{clip}}(\theta,t)$ at position $t$ is defined as \begin{equation}\label{eq:l_clip}\mathcal{L}_{\text{clip}}(\theta,t) = \min\Bigl( r_{i,t}(\theta)\hat{A}_{i,t}, \operatorname{clip}(r_{i,t}(\theta), 1-\epsilon, 1+\epsilon)\hat{A}_{i,t} \Bigr),\end{equation}
and the KL regularization  $\mathcal{D}_{\text{KL}}(\theta,t)$ against the reference policy $\pi_{\text{ref}}$ is 
\begin{equation}\label{eq:d_kl}
\mathcal{D}_{\text{KL}}(\theta,t) = \text{KL} \Bigl[ \pi_{\theta}(\cdot \mid x, y_{i,<t}) \Big|\Big| \pi_{\text{ref}}(\cdot \mid x, y_{i,<t}) \Bigr].
\end{equation}

Here, $r_{i,t}(\theta)$ denotes the probability ratio between the current and old policies, and $y_{i,<t}$ denotes tokens sampled up to the decoding step $t$. Hyperparameters $\epsilon$ and $\beta$ control the clipping range and the KL-regularization strength, respectively. The advantage $\hat{A}_{i,t}$ is calculated based on the rewards within the group. For agentic search, a token mask $I(y_{i,t})$ is applied to ensure that the model does not learn to reproduce or memorize the outputs obtained from external retrieval tools.



\subsection{Temporal Logic}
\Ac{TL} is a formal language for describing how events evolve over time using logical and temporal operators. We illustrate temporal logic given the following query and answer: \textit{``After the woman pours hot water and spoons yogurt into the bowl, what does she place as a topping? Answer: Strawberry''}. Events are represented as atomic primitives that evaluate to \texttt{True} or \texttt{False} at each time step. These primitives are combined with the logical operators \texttt{AND} ($\wedge$), \texttt{OR} ($\vee$), \texttt{NOT} ($\neg$) and \texttt{IMPLY} ($\rightarrow$), and the temporal operators \texttt{ALWAYS} ($\Box$), \texttt{EVENTUALLY} ($\diamondsuit$), \texttt{NEXT} ($\bigcirc$) and \texttt{UNTIL} ($\mathcal{U}$).

Returning to our running example, we define a set of atomic primitives $\mathcal{P}$ and a corresponding \ac{TL} specification $\varphi$ as follows:
\begin{equation*}
    \label{eq:eg_spec}
    \begin{aligned}
        \mathcal{P} &= \{ \texttt{woman}, \; \texttt{pours water}, \; \texttt{spoons yogurt}, \; \texttt{adds strawberry} \}, \\
        \varphi &= \Box\;(\texttt{woman}) \wedge  \left( (\texttt{pours water}  \wedge \texttt{spoons yogurt}) \wedge \diamondsuit \; (\texttt{adds strawberry}) \right).
    \end{aligned}
\end{equation*}
This specification encodes the requirements of the context that \texttt{VSeek} would need to build. This allows us to verify whether the video context retrieves the correct evidence, providing a checkable objective for reinforcement learning.

\section{Method}

In this section, we introduce \texttt{VSeek}, an agentic framework designed to perform precise evidence seeking within long-form videos. We first formalize this interaction as a sequential decision process (\S\ref{sec:method:trajectory}). We then detail our tool infrastructure, which leverages both semantic visual embeddings and subtitle-based indexing to bridge the gap between natural language queries and temporal video segments (\S\ref{sec:method:tools}). Finally, to overcome the challenge of sparse rewards in reinforcement learning for video tasks, we present \ac{VETL}, a neuro-symbolic reward shaping mechanism that provides dense, step-wise feedback based on the retrieval of verifiable visual primitives (\S\ref{sec:method:vetl_rewards}).

\subsection{Interactive Video-Evidence Seeking Trajectory}
\label{sec:method:trajectory}
Following agentic search frameworks in \acp{LLM}, we formulate \texttt{VSeek} as a sequential decision-making process where a single \ac{VLM} interacts with an environment via retriever-based tools. 

Given an initial user prompt $x$ and a long video $V$, the initial context of the agent is defined as $c_0 = x$. At each turn $j$, we autoregressively decode reasoning tokens $a^{\text{think}}_j$ (within \texttt{<think>} and \texttt{</think>}), followed by a search action $a_j$ formatted as \texttt{<search><args>...}\texttt{</args></search>} from the \ac{VLM} policy $\pi_\theta$. Therefore, the generated response $y_j$ at turn $j$ is $ (a^{\text{think}}_j, a_j)  \sim \pi_\theta(\cdot \mid c_{j-1})$ conditioned on an aggregated context $c_{j-1}$.  Executing the search action $a_j$ with the search engine yields a specific segment of frames $o_j \subset V$ retrieved from the source video $V$. The agent's context is then sequentially augmented with the generated tokens and the retrieved video clip as $c_j = [c_{j-1}, a^{\text{think}}_j, a_j, o_j]$.
The agent's trajectory terminates at turn  $J$ when the \ac{VLM} emits its final response within \texttt{<answer>} and \texttt{</answer>}. This yields a complete trajectory $\tau = (c_0, a^{\text{think}}_1, a_1, o_1, \dots, a^{\text{think}}_J, a_J)$. Upon termination, the environment evaluates the response and assigns a terminal exact match reward $r_J \in \{0, 1\}$ (also referred to as $R_{\text{em}}$) based on the final accuracy of the \ac{VQA}. During post-training, the objective is to optimize $\pi_\theta$ to maximize the expected reward. 

\subsection{Agentic Design and Tool Infrastructure}
\label{sec:method:tools}
We built \texttt{VSeek} around a \ac{VLM} configured to seamlessly alternate between reasoning and evidence acquisition. A strict design constraint is that all tool outputs must fit within the \ac{VLM}'s visual context window; therefore, we enforce a per-turn budget of exactly 16 frames for observation $o_t$.

\paragraph{Preprocessing and Indexing.}
Before agent interaction, we preprocess the prescribed long video $V$ by segmenting them into 8-second fixed-length clips. Each clip is embedded using the ViCLIP \cite{wang2023internvid} video encoder referenced as $e_{\text{video}}$, whereas language queries and subtitles are embedded with ViCLIP's text encoder $e_{\text{text}}$. Although alternative temporal grounding models such as CLIP \cite{radford2021learning}, LanguageBind \cite{zhu2023languagebind}, GroundingDino \cite{groundingdino}, or other event detection models \cite{lstm-event-detect,action-detect,event-detect-spatiotemporal} can, in principle, be employed, we adopt ViCLIP in this work owing to its demonstrated performance and inference efficiency. Concurrently, we index the video's subtitles, explicitly mapping each text transcript to the exact frame windows of its occurrence. This dual-index forms the foundation for the agent's retrieval mechanisms.

\paragraph{Actionable Tools.}
Following indexing, we define a suite of tools that allow the \ac{VLM} to query the video index solely via natural language. The model invokes these tools using explicit XML-style structural tags:

\begin{enumerate}
    \item \textbf{Search-based Tool:} Invoked via the \texttt{<search> search\_string </search>} tag, this tool performs semantic visual retrieval by embedding the query with the ViCLIP text encoder and scoring it against pre-computed video clip embeddings. It returns the top-$k$ clips $\mathcal{K}$ that maximize cosine similarity:
\begin{equation}
\mathcal{K} = \mathop{\text{Top-}k}_{v_i \in V} \left( \frac{e_{\text{text}}(\Tilde{q}) \cdot e_{\text{video}}(v_i)}{\|e_{\text{text}}(\Tilde{q})\| \|e_{\text{video}}(v_i)\|} \right),
\end{equation}
where $\Tilde{q}$ is the \texttt{search\_string}, and $v_i$ is the $i$-th clip in the index.  To satisfy the constraint of the visual context, the environment uniformly samples frames from clips in $\mathcal{K}$ to form the observation $o_t$, capped at 16 frames.
    
    \item \textbf{Subtitle-based Search:} This tool searches the video index by subtitle using the \texttt{<search\_subtitle> search\_string </search\_subtitle>} tag. It bypasses the visual embedding space and directly matches the text query $\Tilde{q}$ to the indexed transcripts via
\begin{equation}
    s^* = \mathop{\arg\max}_{s_i \in S} \left( \frac{e_{\text{text}}(\Tilde{q}) \cdot e_{\text{text}}(s_i)}{\lVert e_{\text{text}}(\Tilde{q}) \rVert \, \lVert e_{\text{text}}(s_i) \rVert} \right)
\end{equation}
where $S$ is the set of all subtitles in the video. Using the temporal window of the matched subtitle $s^*$, it returns the corresponding frames of the video clip $v^*$ as $o_t$, enabling precise retrieval of specific spoken dialog or named entities that semantic visual embeddings miss.
    
    \item \textbf{Video Summary:} Invoked via the \texttt{<summary></summary>} tag, this tool gives a global view of the video. When the agent lacks priors for a targeted search, it can trigger this exploratory action. The environment skips the search index and returns 16 frames uniformly sampled throughout the duration of $V$, anchoring the agent and providing the temporal context needed to formulate more precise search queries later.
\end{enumerate}

\begin{figure}[t]
    \centering
    \includegraphics[width=0.8\linewidth]{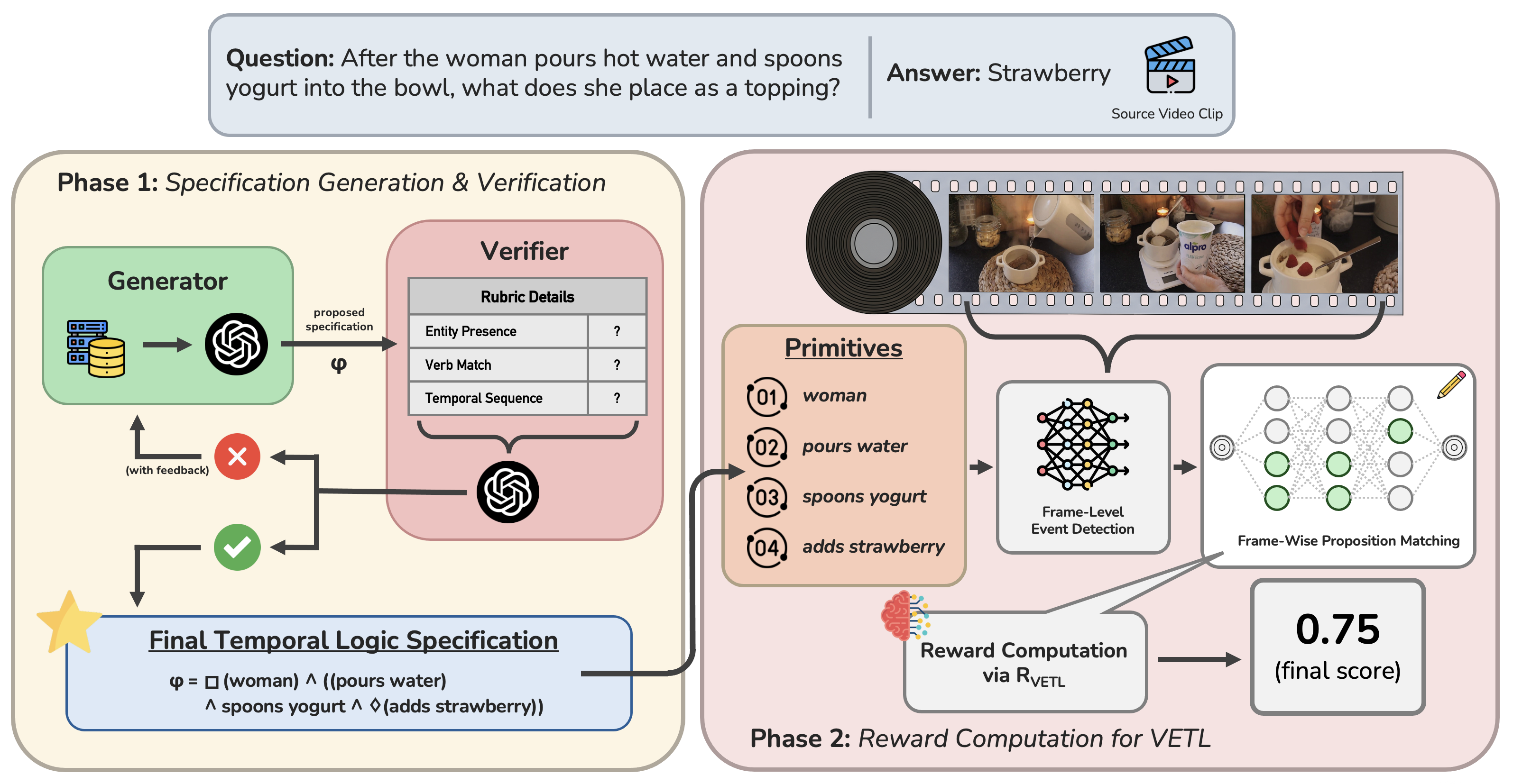}
    \caption{\textbf{VETL Rewards.} We separately compile the question and its correct answer into a temporal logic specification in Phase 1. We utilize the extracted primitives from the temporal logic specification to verify whether the agent's retrieved video context would satisfy answering the question in Phase 2.}
    \label{fig:running_example_vetl}
\end{figure}

\subsection{VETL: Visual Evidence via Temporal Logic}
\label{sec:method:vetl_rewards}

We introduce \ac{VETL}, a neuro-symbolic framework that compiles a natural language question $q$ and reference answer $a$ into a \ac{TL} specification, enabling dense rewards over $\tau$.

\paragraph{Specification Generation.}
\Ac{VETL} derives reliable temporal-logic specifications via an iterative \emph{generate--verify} loop (Algorithm~\ref{alg:vetl_gen}). An \ac{LLM} proposes a \ac{TL} specification $\varphi$ over a set of atomic primitives $\mathcal{P}$ (\textit{e.g.}, object-centric predicates such as \texttt{car} or event-centric predicates such as \texttt{turn-left}), following prior work \cite{sharan2025neuro,shah2025neus} [Line 3]. A judge model then scores $\varphi$ under strict rubrics (in the Appendix) for entity coverage and temporal faithfulness [Line 4]. The judge also returns structured feedback to the generator, and the process repeats until the score is above the threshold $\kappa$ set to 5 [Line 5].

\paragraph{Reward Composition.}


Once the formal specification $\varphi$ is accepted, the agent's visual observations $(o_1,\dots,o_T)$ are mapped to a dense reward. This reward incentivizes the agent to surface the specific visual evidence mandated by the specification. Let $\mathcal{P}_{\varphi}$ denote the finite set of unique visual primitives extracted from $\varphi$ (see Figure \ref{fig:running_example_vetl}). We define a deterministic visual checker as a detection function using a Deep Learning Model (ViCLIP in our case) $D_p(o_t) \in \{0,1\}$, which evaluates to $1$ if primitive $p \in \mathcal{P}_{\varphi}$ is present in observation $o_t$, and $0$ otherwise. The reward is the fraction of required primitives successfully observed across the entire retrieved video context:
\begin{equation}
    R_{\text{VETL}}(\tau; \varphi) = \frac{1}{|\mathcal{P}_{\varphi}|} \sum_{p \in \mathcal{P}_{\varphi}} \max_{1 \le t \le T} D_p(o_t)
    \label{eq:vetl_primitive_reward}
\end{equation}

Notably, while $\varphi$ is formalized in temporal logic \cite{Baier2008,Temporal-and-Modal-Logic}, we use temporal operators only to disentangle semantics from their temporal relationships. Accordingly, we ignore the order in which evidence is retrieved, since the agent's search is non-chronological (\textit{e.g.}, it may search for a later event in the prompt, first). The final reward is given by $R = \lambda_1 R_{\text{em}} + \lambda_2 R_{\text{VETL}}$. Here, $\lambda_1 + \lambda_2 = 1$, are set to 0.7 and 0.3 respectively, and $R_{\text{em}}$ is the exact match reward.

{
\setlength{\textfloatsep}{6pt}
\setlength{\intextsep}{6pt}
\setlength{\floatsep}{6pt}
\setlength{\abovecaptionskip}{2pt}
\setlength{\belowcaptionskip}{2pt}
\begin{algorithm}[t]
\caption{\ac{VETL} Phase 1: Neuro-Symbolic Specification Generation}
\label{alg:vetl_gen}
\begin{algorithmic}[1]
\REQUIRE Question $q$, reference answer $a$ (optional), score threshold for acceptance $\kappa$, max rounds $N$, judge rubrics $\mathcal{R}$, LLM-based temporal specification generator \textsc{Generator}, and LLM-based Judge \textsc{Judge} .
\ENSURE Accepted TL specification $\varphi$.

\STATE $\textit{fb} \leftarrow \emptyset$ \COMMENT{initialize structured feedback}
\FOR{$r=1$ to $N$}
    \STATE $\varphi \leftarrow \textsc{Generator}(q, a, \textit{fb})$ \COMMENT{propose TL over primitives $\mathcal{P}$}
    \STATE $(\textit{score}, \textit{fb}) \leftarrow \textsc{Judge}(q, a, \varphi; \mathcal{R})$ \COMMENT{evaluate against rubrics}
    \IF{$\textit{score} \geq \kappa$}
        \STATE \textbf{break}
    \ENDIF
\ENDFOR

\RETURN $\varphi$
\end{algorithmic}

\end{algorithm}
}

\section{Experiments}
\label{sec:experiments}

This section evaluates \texttt{VSeek} with a focus on verified-signal rewards for post-training and long-horizon video understanding with tool use.

\subsection{Research Questions}
\label{sec:exp:rqs}
We structure our experimental evaluation around four research questions:
\begin{enumerate}
    \item \textbf{RQ1 (Reward Effectiveness):} How does a standard language-based search agent compare to one incentivized by our verified-signal reward?
    \item \textbf{RQ2 (Long-Video Understanding):} Does the reward mechanism yield a significant advantage as video duration increases?
    \item \textbf{RQ3 (Frame Efficiency):} What efficiency gains arise in computational overhead and average frames processed per query?
    \item \textbf{RQ4 (Algorithmic Robustness):} Does adding a verified reward consistently improve performance across multiple post-training \ac{RL} algorithms?
\end{enumerate}
\subsection{Experimental Setup}
\label{sec:exp:setup}

\paragraph{Model Zoo and Baselines.} To evaluate the efficacy of the proposed framework, we benchmark against a diverse set of multimodal architectures categorized into static baselines and active agentic models. Our inference-only baselines include frontier models such as \texttt{GPT-5.2}, and \texttt{InternVL-3.5-4B}, along with Qwen3 variants such as \texttt{Qwen3-VL-4B-Thinking} and  \texttt{Qwen3-VL-4B-Instruct}. For agentic evaluation, we compare our performance with VideoTree \cite{wang2025videotree}, whose captioning and \ac{QA} frameworks are constrained on the \texttt{Qwen3-VL-4B-Instruct} model, and specialized variants of our framework, including Naive-RAG, \texttt{VSeek-GPT}, \texttt{VSeek-EM}, and \texttt{VSeek-VETL}. Naive-RAG retrieves the top 64 frames that have the highest similarity with respect to the original question, whereas \texttt{VSeek-GPT} performs the same natural language-based search with \texttt{GPT-5.2} as the base model. Both \texttt{VSeek-EM} and \texttt{VSeek-VETL} are post-trained specifically on top of \texttt{Qwen3-VL-4B-Thinking} models to leverage their native chain-of-thought capabilities for complex video search trajectories. \texttt{VSeek-EM} is post-trained with \ac{GRPO} with the exact match reward, whereas \texttt{VSeek-VETL} incorporates the \ac{VETL} reward defined in Equation \ref{eq:vetl_primitive_reward}.

\paragraph{Video Indexing and Retrieval.} A core component of our architecture is the structured representation of raw video streams. We use ViCLIP to generate dense embeddings over segmented videos, indexing each video with fixed 8-second segments. This granularity lets the retriever efficiently align natural language queries with specific video clips and enables a hierarchical coarse-to-fine search strategy. Using the same 8-second indexing across all benchmarks ensures consistent event grounding for the agent’s search actions.

\paragraph{Datasets and Benchmarks.} Our evaluation covers both \ac{ID} and \ac{OOD} settings to probe whether the \texttt{VSeek} variants remain performant and reliable across multiple genres of questions. In-domain performance is measured on LongVideoBench \cite{wu2024longvideobench}, MLVU \cite{zhou2025mlvu}, and VideoMME \cite{fu2025video}, which require accurate retrieval over long videos. To assess zero-shot transfer, we also report results on CGBench-mini \cite{chen2024cg} and LVBench \cite{wang2024lvbench}, challenging \ac{OOD} benchmarks for testing the robustness of language-based search on unseen video distributions.

\paragraph{Evaluation Metrics.} We evaluate performance using metrics based on success and efficiency. We measured the \textit{Pass@1} and \textit{Pass@4}, which capture the probability of generating one correct answer across 1 and 4 passes, respectively, and \textit{Majority@4}, which assesses consensus accuracy across 4 independent samples. For efficiency, we report the average number of frames sampled per video, indicating how selectively the model searches for sparse visual evidence and the resulting computational cost for long-video understanding.

\paragraph{Implementation Details.} Full hyperparameters and training prompts are in the Appendix. We train on 8 GH200s for 900 steps with a batch size of 32.

\subsection{RQ1: Performance with the Verified-Signal Reward}
\label{sec:exp:rq1}

\newcommand{\gain}[2]{\shortstack{#1\\[-0.3ex]{\scriptsize\textcolor{green!60!black}{#2}}}}
\newcommand{\drop}[2]{\shortstack{#1\\[-0.3ex]{\scriptsize\textcolor{red!70!black}{#2}}}}

        \begin{table*}[t]
            \centering
            \small
            \setlength{\tabcolsep}{3pt}
            \caption{Results on three ID (*) and two OOD benchmarks. \texttt{VSeek} is initialized from \texttt{Qwen3-VL-4B-Thinking}; for the post-trained variants \texttt{VSeek-EM} and \texttt{VSeek-VETL}, we report the \textit{absolute score} (best in \textbf{bold} and second best in \underline{underline}) and the \textit{relative gain} (\textcolor{green}{+}/\textcolor{red}{-}\%) with respect to direct inference on the base model (\texttt{Qwen3-VL-4B-Thinking}). Note that \texttt{VSeek-EM} is trained with an exact-match reward only, while \texttt{VSeek-VETL} uses \ac{VETL} rewards. Additionally, we report the performance of the \texttt{GPT-5.2} variant of \texttt{VSeek} (\texttt{VSeek-GPT}).}
            
            \resizebox{\textwidth}{!}{
                \begin{tabular}{l|ccc|ccc|ccc|ccc|ccc}
                    \hline
                    \multirow{2}{*}{\textbf{Method}} & \multicolumn{3}{c|}{\textbf{LongVideoBench}\textsuperscript{*}} & \multicolumn{3}{c|}{\textbf{Video-MME}\textsuperscript{*}} & \multicolumn{3}{c|}{\textbf{MLVU}\textsuperscript{*}} & \multicolumn{3}{c|}{\textbf{LVBench}} & \multicolumn{3}{c}{\textbf{CGBench-mini}} \\
                     & p@1 & p@4 & m@4 & p@1 & p@4 & m@4 & p@1 & p@4 & m@4 & p@1 & p@4 & m@4 & p@1 & p@4 & m@4 \\
                    \hline
                    \rowcolor[HTML]{F2F2F2} \multicolumn{16}{l}{\textit{Proprietary Models}} \\
                    \hline
                    \texttt{GPT-5.2} (64 frames) & 57.8 & 63.5 & 58.1 & \textbf{67.7} & 72.9 & \textbf{67.9} & 61.3 & 67.1 & 61.5 & 38.8 & 46.1 & 39.1 & \textbf{36.7} & 44.3 & \underline{37.2} \\
                    \texttt{VSeek-GPT}  & 54.4 & \textbf{72.8} & 57.2 & 61.2 & \textbf{77.8} & \underline{65.3} & 56.3 & 75.4 & 59.9 & \textbf{40.3} & \textbf{58.5} & \underline{40.9} & 34.4 & \textbf{56.2} & \textbf{39.9} \\
                    
                    \hline
                    \rowcolor[HTML]{F2F2F2} \multicolumn{16}{l}{\textit{Open Source Models}} \\
                    \hline
                    \texttt{InternVL-3.5-4B}  & 38.9 & 60.4 & 42.9 & 53.1 & \underline{76.1} & 54.7 & 41.8 & 69.1 & 43.4 & 32.6 & \underline{57.7} & 33.2 & 21.5 & 49.2 & 22.3 \\
                    \texttt{Qwen3-VL-4B-Thinking} & 51.7 & 60.8 & 53.7 & 58.1 & 65.2 & 60.0 & 58.0 & 64.3 & 58.8 & 35.6 & 41.8 & 36.4 & 31.6 & 41.6 & 32.4 \\
                    \texttt{Qwen3-VL-4B-Instruct} & 54.4 & 59.2 & 54.9 & \underline{63.7} & 68.6 & 64.3 & 61.9 & 66.2 & 62.4 & 38.7 & 47.7 & 38.6 & \underline{34.7} & 43.5 & 34.7 \\
                    \texttt{Qwen3-VL-8B-Thinking} & 51.7 & 68.6 & 56.3 & 60.3 & 73.4 & 63.0 & 56.3 & 71.5 & 59.4 & 35.1 & 51.3 & 37.4 & 31.3 & 49.9 & 33.7 \\
                    \texttt{Qwen3-VL-8B-Instruct} & 54.5 & 59.8 & 54.9 & \underline{63.7} & 68.6 & 64.3 & 61.9 & 66.2 & 62.4 & 38.7 & 47.7 & 38.6  & \underline{34.7} & 43.5 & 34.7 \\

                    \hline
                    \rowcolor[HTML]{F2F2F2} \multicolumn{16}{l}{\textit{Agentic Systems}} \\
                    \hline
                    Naive RAG  & 49.8 & 60.9 & 49.8 & 55.0  & 63.3 & 56.5 & 55.6 & 64.0 & 57.5 & 37.1 & 46.1 & 38.7 & 30.4 & 41.2 & 32.2 \\
                    VideoTree  & 45.4 & 47.6 & 45.1 & 56.3  & 58.0 & 56.6 & 46.1 & 48.2 & 46.8 & 34.0 & 36.4 & 34.2 & 29.6 & 31.9 & 30.0 \\
                    \texttt{VSeek-EM} & \gain{\underline{60.5}}{+8.8\%} & \gain{\underline{71.1}}{+10.3\%} & \gain{\underline{62.1}}{+8.4\%} & \gain{59.3}{+1.2\%} & \gain{71.2}{+6.0\%} & \gain{60.1}{+0.1\%} & \gain{\underline{66.7}}{+8.7\%} & \gain{\underline{75.9}}{+11.6\%} & \gain{\underline{67.3}}{+8.5\%} & \gain{\underline{40.0}}{+4.4\%} & \gain{54.9}{+13.1\%} & \gain{\textbf{41.5}}{+5.1\%} & \gain{33.2}{+1.6\%} & \gain{\underline{52.3}}{+10.7\%} & \gain{34.5}{+2.1\%} \\
                    \texttt{VSeek-VETL}  & \gain{\textbf{60.9}}{+9.2\%} & \gain{70.7}{+9.9\%} & \gain{\textbf{62.6}}{+8.9\%} & \gain{61.4}{+3.3\%} & \gain{72.0}{+6.8\%} & \gain{62.8}{+2.7\%} & \gain{\textbf{67.9}}{+9.9\%} & \gain{\textbf{77.6}}{+13.3\%} & \gain{\textbf{68.8}}{+10.0\%} & \gain{38.8}{+3.2\%} & \gain{53.0}{+11.2\%} & \gain{39.7}{+3.3\%} & \gain{32.3}{+0.7\%} & \gain{51.1}{+9.5\%} & \gain{33.3}{+0.9\%} \\
                    
                    \hline
                \end{tabular}
            }
            \label{tab:rq1_results}
        \end{table*}

\paragraph{Performance of Agentic Search.} As shown in Table~\ref{tab:rq1_results}, the transition from passive perception to an active, natural language-driven search policy yields significant gains over existing video agent baselines. By utilizing a flexible language interface for evidence seeking, \texttt{VSeek} effectively pinpoints sparse evidence. \texttt{VSeek} variants consistently outperform other agentic systems like Naive RAG, with the most substantial improvements observed on \ac{ID} datasets. While gains on \ac{OOD} benchmarks are more modest, they remain positive. We expect that larger base models will further enhance the generalizability of this search policy, leading to more sophisticated cross-domain evidence acquisition.

\paragraph{Impact of \ac{VETL} Rewards.}
The introduction of neuro-symbolic verified rewards from \ac{VETL} provides a distinct advantage over \ac{EM} baselines by enforcing evidence coverage during video context construction. On in-domain evaluations, this verification signal drives a performance increase, specifically a 1.2\% gain on MLVU for \texttt{VSeek-VETL} over \texttt{VSeek-EM}, as the model is explicitly rewarded for evidence seeking.

In contrast, we observe a modest degradation in Pass@1 scores compared to the agent trained with exact match reward on \ac{OOD} evaluations.  Error analysis suggests that this drop stems from the high specificity of learned search strings, which can become overly specialized to the in-distribution corpus. We believe that this is a ``specialization bottleneck'' rather than a fundamental flaw; a more diverse training corpus should allow the model to learn generalized search primitives and strategies that maintain high performance across unseen genres without sacrificing the rigor of the verification signal.

\begin{table*}[t]
    \centering
    \small
    \setlength{\tabcolsep}{3pt}
    \caption{\textbf{RQ2} results: Performance (best in \textbf{bold} and second best in \underline{underline}) of post-trained \texttt{VSeek} variants with gains (\textcolor{green}{+}/\textcolor{red}{-}\%) reported relative to the \texttt{Qwen3-VL-4B-Thinking} base model. We include \texttt{VSeek-GPT} (utilizing \texttt{GPT-5.2} for reasoning) as a strong agentic baseline. Our post-trained models consistently outperform this variant. Additionally, we show that, particularly on longer videos ($>$10m on LongVideoBench and $>$1m in Video-MME and MLVU), the models post-trained with the verified-signal reward have better performance than the model variant post-trained with only the exact-match reward.}
    \resizebox{\textwidth}{!}{%
        \begin{tabular}{l|ccc|ccc|ccc}
            \hline
            \multirow{2}{*}{\textbf{Method}} & \multicolumn{3}{c|}{\textbf{LongVideoBench}} & \multicolumn{3}{c|}{\textbf{Video-MME}} & \multicolumn{3}{c}{\textbf{MLVU}} \\
             & $<$1m & 1--10m & 10--60m & $<$1m & 1--10m & 10--60m & 1--10m & 10--60m & 60m+ \\
            
            \hline
            \rowcolor[HTML]{F2F2F2} \multicolumn{10}{l}{\textit{Baselines}} \\
            
            \hline
            \texttt{Qwen3-VL-4B-Instruct} & 62.5 & 56.8 & 48.9 & 68.1 & \textbf{69.8} & 57.7 & 63.4 & 59.9 & 51.2 \\
            \texttt{Qwen3-VL-4B-Thinking} & 60.9 & 56.3 & 47.9 & 69.4 & 64.7 & 53.9 & 58.9 & 59.9 & 51.2 \\
            Naive RAG          & 58.1 & 55.5 & 46.2 & 56.8 & 60.9 & 52.2 & 56.8 & 60.8 & 61.2 \\
            VideoTree          & 40.2   & 44.4   & 40.8   & 61.0   & 58.5   & 53.8   & 44.7   & 55.3   & 40.5   \\
            
            \hline
            \rowcolor[HTML]{F2F2F2} \multicolumn{10}{l}{\textit{Post-trained Variants}} \\
            
            \hline
            \texttt{VSeek-EM}   & \gain{69.1}{+8.2\%} & \textbf{\gain{62.8}{+6.5\%}} & \gain{57.5}{+9.6\%} & \drop{66.7}{-2.7\%} & \gain{65.0}{+0.3\%} & \drop{53.8}{-0.1\%} & \gain{67.0}{+8.1\%} & \gain{69.0}{+9.1\%} & \textbf{\gain{62.2}{+11.0}} \\
            \texttt{VSeek-VETL} & \textbf{\gain{69.6}{+8.7\%}} & \gain{61.9}{+5.6\%} & \textbf{\gain{58.9}{+11.0\%}} & \drop{68.6}{-0.8\%} & \gain{66.6}{+1.9\%} & \gain{57.7}{+3.8\%} & \textbf{\gain{68.8}{+9.9\%}} & \textbf{\gain{69.9}{+10.0\%}} & \textbf{\gain{62.2}{+11.0\%}} \\
            
            \hline
            \rowcolor[HTML]{F2F2F2} \multicolumn{10}{l}{\textit{Proprietary Models}} \\
            
            \hline
            \texttt{VSeek-GPT}  & 59.5 & 59.1 & 54.4 & \textbf{69.5} & 68.9 & \textbf{60.6} & 59.2 & 63.1 & 55.8 \\
            \hline
        \end{tabular}%
    }
    \label{tab:rq2_results}
\end{table*}

\subsection{RQ2: Long Video Understanding with Tool Use}
\label{sec:exp:rq2}

Our results show that both \texttt{VSeek-VETL} and \texttt{VSeek-EM} maintain substantially higher reasoning stability than baseline agentic systems as video length increases. While methods like VideoTree and Naive RAG preprocess and select scenes of interest separately from reasoning and answering, both \texttt{VSeek} variants interleave search directly with thinking and response generation. This integration enables more coherent answers on longer videos, as the model can dynamically adapt its search strategy based on intermediate reasoning and gathered evidence. Moreover, by awarding partial credit for verifiable intermediate steps via our neuro-symbolic pipeline, the verified reward provides a denser training signal than a binary success/failure metric, encouraging the model to first localize key events and rigorously verify them before synthesizing an answer.

\begin{figure}[t]
    \centering
    \includegraphics[width=\linewidth]{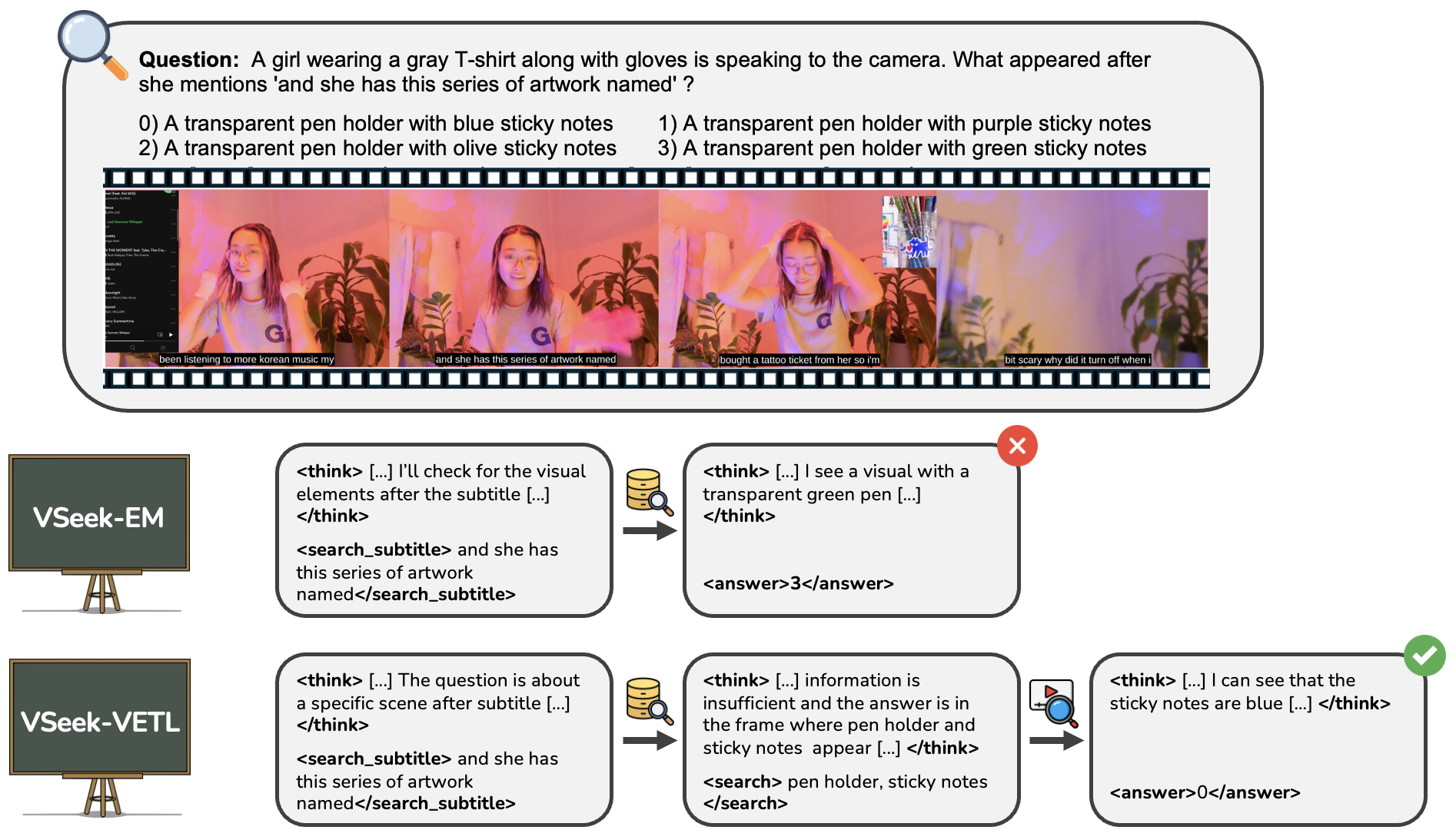}
\caption{ \textbf{Qualitative analysis.}  This example qualitatively highlights the advantage of \texttt{VSeek-VETL} over outcome-only based training. Here, both \texttt{VSeek-EM} and \texttt{VSeek-VETL} correctly decide that the first clip to retrieve must contain the desired subtitle. In contrast to \texttt{VSeek-EM}, which immediately hallucinates an answer, \texttt{VSeek-VETL} devises a more targeted search query to retrieve a set of segments corresponding to a pen holder to produce the correct answer.}
    \label{fig:qualitative-fig}
\end{figure}

In longer videos, \texttt{VSeek-VETL} outperforms \texttt{VSeek-EM} and VideoTree by providing more specific and accurate grounding. Qualitatively, we observe a shift in evidence-seeking behavior: \texttt{VSeek-EM} often issues generic queries which can retrieve irrelevant clips, whereas \texttt{VSeek-VETL} learns to produce highly specific keywords.  Since these keywords do not directly match the atomic primitives from the \ac{VETL} pipeline, rewards are devised based on the retrieved video context, to ensure that it is closely aligned with the query.  Therefore, this search is enabled by the \ac{VETL}-based reward, which trains the model to explore better search strings rather than repeatedly issuing unproductive search queries, a common failure of baseline agentic systems. Figure~\ref{fig:qualitative-fig} shows how \texttt{VSeek-VETL} formulates more targeted queries to navigate a long video.

Because both \texttt{VSeek-VETL} and \texttt{VSeek-EM}  are post-trained to answer questions with the search-based tool, both methods learn a precise internal model of the video's temporal structure, allowing them to find sparse evidence in hour-long raw footage with highly targeted search actions. This post-training yields better reasoning capability, even outperforming \texttt{VSeek-GPT}, which uses \texttt{GPT-5.2} as the reasoning model to issue search queries and answer questions.

\subsection{RQ3: Efficiency Gains}
\label{sec:exp:rq3}

We evaluate \texttt{VSeek}'s computational efficiency via average frames processed per query. Standard \acp{VLM} and \ac{RAG} baselines typically use a fixed 64-frame budget, whereas \texttt{VSeek} variants achieve better accuracy with under 20 frames on average. For VideoTree, which requires frame-level captions and samples these captions to adjudicate query relevancy for \ac{QA}, we report the number of sampled captions.  As shown in Table~\ref{tab:efficiency_frames_expanded}, \texttt{VSeek-VETL} samples slightly more frames than \texttt{VSeek-EM} (\textit{e.g.}, $31.9$ vs. $30.3$ on LongVideoBench) due to the evidence-based reward where penalizing unverified claims encourages more thorough retrieval. This trade-off ensures that the retrieved context meets the temporal and causal requirements of the query, improving the quality of the reasoning. \texttt{VSeek-VETL} however does not outperform much larger proprietary systems like \texttt{VSeek-GPT}, which averages 24--29 frames on long-form content. Coupled with the degradation of the performance on these tasks, we observe that \texttt{GPT-5.2} often answers questions without complete information, resulting in the lower frame usage.

        \begin{table*}[t]
            \centering
            \small
            \setlength{\tabcolsep}{4pt}
            \caption{Efficiency Analysis (\textbf{RQ3}): Comparison of the average frames sampled across diverse video durations for three in-domain datasets. \texttt{VSeek-VETL} significantly reduces the visual footprint while achieving state-of-the-art performance.}
            \resizebox{\textwidth}{!}{
                \begin{tabular}{l | cccc | cccc | cccc}
                    \hline
                    \multirow{2}{*}{\textbf{Method}} & \multicolumn{4}{c|}{\textbf{LongVideoBench}} & \multicolumn{4}{c|}{\textbf{Video-MME}} & \multicolumn{4}{c}{\textbf{MLVU}} \\
                     &$<$1m & 1m-10m & 10m-60m & \textbf{Avg.} & $<$1m & 1m-10m & 10m-60m & \textbf{Avg.} & 1-10m & 10m-60m & 60m+ & \textbf{Avg.} \\
                    
                    \hline
                    \rowcolor[HTML]{F2F2F2} \multicolumn{13}{l}{\textit{Baselines}} \\
                    
                    \hline
                    \texttt{Qwen3-VL-4B-Instruct}  & 64.0 & 64.0 & 64.0 & 64.0 & 64.0 & 64.0 & 64.0 & 64.0 & 64.0 & 64.0 & 64.0 & 64.0 \\
                    \texttt{Qwen3-VL-4B-Thinking}  & 64.0 & 64.0 & 64.0 & 64.0 & 64.0 & 64.0 & 64.0 & 64.0 & 64.0 & 64.0 & 64.0 & 64.0 \\
                    Naive RAG         & 64.0 & 64.0 & 64.0 & 64.0 & 64.0 & 64.0 & 64.0 & 64.0 & 64.0 & 64.0 & 64.0 & 64.0 \\
                    VideoTree         & 13.6   & 206.7   & 384.3   & 232.6  & 26.5   & 111.3   & 256.6   & 168.8   & 213.7   & 323.5   & 402.4   & 240.7   \\
                    
                    \hline
                    \rowcolor[HTML]{F2F2F2} \multicolumn{13}{l}{\textit{VSeek Variants}} \\
                    
                    \hline
                    \texttt{VSeek-EM}  & 30.3 & 31.2 & 31.4 & 31.0 & 29.5 & 30.2 & 31.6  & 30.8 &  30.1 & 30.4 & 30.5 & 30.2 \\
                    \texttt{VSeek-VETL} & 31.9 & 31.9 & 32.1 & 32.0 & 31.8 & 31.9 & 32.0  & 31.9 & 31.9 & 31.6 & 31.3 & 31.8 \\
                    
                    \hline
                    \rowcolor[HTML]{F2F2F2} \multicolumn{13}{l}{\textit{Proprietary Models}} \\
                    
                    \hline
                    \texttt{VSeek-GPT}  & 17.6 & 27.1 & 26.9 & 24.5 & 20.4 & 25.7 & 32.9 & 28.5 & 23.9 & 23.5 & 30.1 & 28.9 \\
                    \hline
                \end{tabular}
            }
            \label{tab:efficiency_frames_expanded}
        \end{table*}
    
\subsection{RQ4: Influence Across Post-Training RL Algorithms}
\label{sec:exp:rq4}

Building on these initial gains, we investigate whether the benefits of verified rewards are consistent across diverse \ac{RL} algorithms. We instantiate \ac{GRPO} \cite{shao2024deepseekmath}, \ac{GSPO} \cite{zheng2025group}, and \ac{RLOO} \cite{ahmadian2024back}, using both exact match reward and our verified reward signal. We show that the \ac{VETL} reward improves performance with \ac{GRPO} and \ac{RLOO} post-training algorithms, while showing degradation with \ac{GSPO}. We provide these results in the Appendix.

\section{Limitations and Future Work}
Despite the performance gains achieved by \texttt{VSeek}, several promising avenues for future research remain.

\begin{enumerate}
\item \textbf{Non-Sequential Search Paradigms}: \texttt{VSeek} currently runs sequentially, which can cause out-of-order frame processing relative to the video timeline. Future work can adopt ``fan-out'' search strategies, similar to deep research frameworks, so that the agent can pursue multiple search strings in parallel.
\item \textbf{Advanced Formal Methods}: While \ac{VETL} is a strong foundation, integrating more expressive temporal logic families can further refine the reward landscape. Especially, in conjunction with better agentic architectures, models can be rewarded better to capture more complex causal and temporal dependencies.
\item \textbf{Retriever and Indexing Bottlenecks}: Video evidence grounding is limited by the underlying retriever and indexing models. Future work can use interleaved search with stronger multimodal retrievers and explore generalization to diverse retrievers.

\end{enumerate}


\label{sec:conclusion}
\section{Conclusion}
In this work, we make two contributions. First, we introduce \texttt{VSeek}, an agentic framework that leverages natural language-based search tools for long video question answering, therefore, moving beyond passive perception and uniform sampling. Then, to ensure that the retrieved context is relevant to the question, we further introduce \ac{VETL}, a neuro-symbolic pipeline from which we derive dense, verified rewards for effective \ac{RL} post-training to link high-level reasoning required for tool calls with low-level visual grounding.

Empirically, \texttt{VSeek} variants consistently outperform static baselines and match strong agentic models like \texttt{GPT-5.2}, especially on longer videos. We also show that our proposed reward \ac{VETL} improves post-training a   nd the final model's performance by ensuring that the agent’s intermediate ``thought-and-search'' gathers relevant video evidence. We hope \texttt{VSeek} and \ac{VETL} offer a solid foundation for future work on stronger video agents and post-training methods for long video understanding.

\section*{Acknowledgements}

This material is based upon work supported in part by the Office of Naval Research (ONR) under Grant No. N00014-22-1-2254. Additionally, this work was supported by the Defense Advanced Research Projects Agency (DARPA) contract DARPA ANSR: RTX CW2231110. Approved for Public Release, Distribution Unlimited. We also like to thank the Texas Advanced Computing Center at The University of Texas at Austin for compute support.

\bibliographystyle{splncs04}
\bibliography{main}

\clearpage
\appendix
\section{Appendix: Extra Results and Discussions}

    \subsection{Performance of Reward on Different RL post-training algorithms}
        Table \ref{tab:appendix_postraining} demonstrates that incorporating the \ac{VETL} reward enables improved performance across a variety of reinforcement learning post-training algorithms when applied to the same base model. Specifically, integrating \ac{VETL} with methods such as \ac{RLOO} yields consistent performance gains over the \ac{EM} baseline on benchmarks including LongVideoBench, Video-MME, and MLVU. However, we observe a notable degradation in performance when utilizing the \ac{GSPO} training algorithm. We believe this deterioration stems from stability challenges and hyperparameter tuning issues, especially for sequence-level ratios, which caused the policy performance to severely degrade during training.
    
        \begin{table*}[ht]
            \centering
            \setlength{\tabcolsep}{3pt}
            \caption{Evaluation of the \ac{VETL} reward across multiple RL post-training algorithms. The table illustrates the performance differences between the proposed \ac{VETL} signal and the exact match (EM) baseline when applied to post-training algorithms such as RLOO and GSPO across five diverse benchmarks.}
                \begin{tabular}{l|c|c|c}
                    \hline
                    \multirow{2}{*}{\textbf{Method}} & \textbf{LongVideoBench} & \textbf{Video-MME} & \textbf{MLVU}   \\
                     & p@1 & p@1 &  p@1  \\
                    \hline
                    \texttt{VSeek-EM} & 60.5 &  59.3 &  66.7  \\
                    \texttt{VSeek-VETL} & \gain{60.9}{+0.4\%}  & \gain{61.4}{+2.1\%} &  \gain{67.9}{+1.2\%}  \\
                    \texttt{VSeek-EM GSPO} & 60.0 &  62.9 & 67.7 \\
                    \texttt{VSeek-VETL GSPO}  & \drop{58.0}{-2.0\%} & \drop{58.1}{-4.8\%} & \drop{62.7}{-5.0\%}  \\
                    \texttt{VSeek-EM RLOO}  & 58.0 & 60.5 &  65.5  \\
                    \texttt{VSeek-VETL RLOO}  & \gain{60.4}{+2.4\%} & \gain{61.3}{+0.8\%} & \gain{66.8}{+1.3\%} \\
                    \hline
                \end{tabular}
            \label{tab:appendix_postraining}
        \end{table*}

\subsection{Significance of Temporal Logic Decomposition}

We aim to disambiguate the contribution of the \ac{TL} layer within our reward pipeline. The \ac{TL} layer explicitly decouples core semantic propositions from complex temporal relationships, ensuring accurate reward assignments during agentic post-training. To isolate and evaluate this contribution, we compare our approach against \texttt{VSeek-Direct}, a variant post-trained using rewards derived from propositions isolated without temporal logic grounding (\textit{i.e.}, direct prompting). We present the prompt utilized for this decomposition in the next paragraph. When utilizing this direct approach, the reward mechanism often fails to enforce semantic propositions are often decoupled from their temporal counterparts or conjoined with each other. 

As demonstrated in Table \ref{tab:reward_ablation}, the \texttt{VSeek-Direct} variant suffers noticeable performance degradations across temporal benchmarks compared to our proposed method. This confirms that the \ac{VETL} formulation, which strictly enforces evidence coverage and temporal sequencing, is crucial for optimal agentic retrieval. Furthermore, independent validation via the \texttt{GPT-5.2} Judge (Table \ref{tab:gpt_judge_reward}) and human evaluations confirm that \ac{VETL} yields significantly higher temporal isolation, semantic alignment, and overall correctness compared to direct decomposition frameworks.

\begin{table*}[t]
    \centering
    \caption{Results of \texttt{VSeek} variants. We analyze 3 post-trained variants: \texttt{VSeek-EM} (exact match reward), \texttt{VSeek-VETL} (proposed reward), and \texttt{VSeek-Direct} (direct decomposition) on \texttt{Qwen3-VL-4B-Thinking}.}
    \label{tab:reward_ablation}
    \vspace{0.25em}
    \resizebox{\textwidth}{!}{%
    \begin{tabular}{l|ccc|ccc|ccc|ccc|ccc}
    \hline
        \multirow{2}{*}{\textbf{Method}} & \multicolumn{3}{c|}{\textbf{LongVideoBench}\textsuperscript{*}} & \multicolumn{3}{c|}{\textbf{Video-MME}\textsuperscript{*}} & \multicolumn{3}{c|}{\textbf{MLVU}\textsuperscript{*}} & \multicolumn{3}{c|}{\textbf{LVBench}} & \multicolumn{3}{c}{\textbf{CGBench-mini}} \\
         & p@1 & p@4 & m@4 & p@1 & p@4 & m@4 & p@1 & p@4 & m@4 & p@1 & p@4 & m@4 & p@1 & p@4 & m@4 \\
        \hline
        \texttt{VSeek-EM} & 60.5 & \textbf{71.1} & 62.1 & 59.3 & 71.2 & 60.1 & 66.7 & 75.9 & 67.3 & \textbf{40.0} & 54.9 & \textbf{41.5} & \textbf{33.2} & \textbf{52.3} & \textbf{34.5} \\
        \texttt{VSeek-VETL}  & \textbf{60.9} & 70.7 & \textbf{62.6} & \textbf{61.4} & 72.0 & \textbf{62.8} & \textbf{67.9} & \textbf{77.6} & \textbf{68.8} & 38.8 & 53.0 & 39.7 & 32.3 & 51.1 & 33.3 \\
        \texttt{VSeek-Direct}  & 59.8 & 70.2 & 61.6 & 59.3 & 70.3 & 60.4 & 66.3 & 76.3 & 67.1 & 39.0 & 53.1 & 39.4 & 31.7 & 50.6 & 32.5 \\
        \hline
    \end{tabular}%
    }
\end{table*}

\begin{table}[t]
    \centering
    \caption{Independent validation of the reward pipeline using the \texttt{GPT-5.2} Judge, averaged across all datasets. Semantic Correctness (Sem.) checks whether propositions are independent of temporal keywords; Temporal Operators (Temp.) verify that temporal relationships are correctly isolated; and Alignment (Align.) measures how well the temporal logic formula matches the question.}
    \label{tab:gpt_judge_reward}
    \vspace{0.25em}
    \scriptsize
    \setlength{\tabcolsep}{3pt}
    \resizebox{0.92\linewidth}{!}{%
    \begin{tabular}{lcccc}
    \toprule
    Reward Type & Mean Score (0-6) & Sem. (0-2) & Temp. (0-2) & Align. (0-2) \\
    \midrule
    \texttt{VSeek-Direct} & 3.8 & 1.3 & 1.2 & 1.3 \\
    \texttt{VSeek-VETL} & 5.3 & 1.8 & 1.8 & 1.7 \\
    \bottomrule
    \end{tabular}%
    }
    \vspace{-0.5em}
\end{table}

\subsection{Analysis of the Instruct Model variants}
We evaluate the agentic post-training pipeline on the instruct-tuned base models, denoted as the \texttt{VSeek-Inst} variants as shown in Table \ref{tab:ins_results}. While Instruct models are directly optimized to follow instructions, we observe that during agentic post-training this underlying bias towards prompt adherence does not help in improving the final performance of the model over their Thinking counterparts.
However,  we do see that the \ac{VETL} reward helps improve performance over models post-trained solely with the Exact Match reward.



\begin{table*}[t]
    \centering
    \caption{Results of \texttt{VSeek} variants. We analyze 2 post-trained variants: EM (exact match reward), and \ac{VETL} (proposed reward), on both Thinking and Instruct (indicated by Ins).}
    \label{tab:ins_results}
    \vspace{0.25em}
    \resizebox{\textwidth}{!}{%
    \begin{tabular}{l|ccc|ccc|ccc|ccc|ccc}
    \hline
        \multirow{2}{*}{\textbf{Method}} & \multicolumn{3}{c|}{\textbf{LongVideoBench}\textsuperscript{*}} & \multicolumn{3}{c|}{\textbf{Video-MME}\textsuperscript{*}} & \multicolumn{3}{c|}{\textbf{MLVU}\textsuperscript{*}} & \multicolumn{3}{c|}{\textbf{LVBench}} & \multicolumn{3}{c}{\textbf{CGBench-mini}} \\
         & p@1 & p@4 & m@4 & p@1 & p@4 & m@4 & p@1 & p@4 & m@4 & p@1 & p@4 & m@4 & p@1 & p@4 & m@4 \\
        \hline
        \texttt{VSeek-EM} & 60.5 & \textbf{71.1} & 62.1 & 59.3 & 71.2 & 60.1 & 66.7 & 75.9 & 67.3 & \textbf{40.0} & 54.9 & \textbf{41.5} & \textbf{33.2} & \textbf{52.3} & \textbf{34.5} \\
        \texttt{VSeek-VETL}  & \textbf{60.9} & 70.7 & \textbf{62.6} & \textbf{61.4} & 72.0 & \textbf{62.8} & \textbf{67.9} & \textbf{77.6} & \textbf{68.8} & 38.8 & 53.0 & 39.7 & 32.3 & 51.1 & 33.3 \\
        \texttt{VSeek-Ins-EM} & 55.4 & 65.2 & 56.1 & 58.7 & 68.9 & 59.9 & 60.3 & 68.5 & 60.8 & 36.0 & 45.8 & 36.9 & 27.2 & 38.1 & 27.5 \\
        \texttt{VSeek-Ins-VETL}  & 55.2 & 67.5 & 57.5 & 60.0 & \textbf{73.3} & 61.7 & 66.1 & 77.1 & 67.6 & 39.4 & \textbf{55.7} & 40.8 & 31.3 & 46.8 & 32.0 \\
        \hline
    \end{tabular}%
    }
\end{table*}

\section{Appendix: Training and Experiment Details}

    \subsection{Experiment Details}
    
        \paragraph{Training Data and Evaluation Protocol.}
        To ensure a rigorous evaluation of the reinforcement learning process, we curate a comprehensive data mixture by randomly splitting the questions and corresponding video samples from the MLVU, VideoMME, and LongVideoBench datasets. Specifically, we allocate 80\% of this mixture for training the policy and reserve the remaining 20\% as an in-domain validation set. Throughout the 900 steps of reinforcement learning training, we continuously evaluate the model on this held-out 20\% split. The final model weights are determined by selecting the checkpoint that achieves the highest validation performance across the entire training trajectory. 
        
        \paragraph{Retriever Configuration.} 
        The search module in \texttt{VSeek} relies on a dedicated retriever server that utilizes videos indexed offline with an embedding model to facilitate retrieval of relevant segments via natural language search queries. First, the raw video corpus is processed offline with the video embedding head of ViCLIP. During this process,  videos are chunked into contiguous 8-second segments and sampled at a strict rate of 1 frame per second (fps), yielding exactly eight frames per clip. The resulting dense embeddings for these discrete clips are stored in a vector database. When the agent formulates a search query, ViCLIP encodes the text into the joint embedding space, retrieving the top four candidate clips based on maximal cosine similarity. To preserve the chronological order crucial for video reasoning models, the retrieval server actively sorts these four independent clips based on their original timestamp indices within the source video. The temporally re-ordered clips are appended to the agent's current working context, providing the necessary visual history to inform subsequent reasoning and search actions.  
        
    \subsection{Visual Evidence via Temporal Logic (VETL) Rewards.}
    
        As previously outlined, we perform an offline decomposition of the provided questions and their corresponding reference answers into verifiable \ac{TL} specifications (Phase 1). The process relies on a lightweight, iterative feedback loop between a generator language model and a judge language model to maximize the fidelity of the extracted temporal and event primitives. Initially, the generator model processes the question and the ground-truth answer text to propose a set of atomic primitives alongside an initial logical formula. This generation strictly isolates independent visual events to prevent the merging of distinct temporal states and restricts the logical operators to a fundamental set comprising conjunction, disjunction, negation, and the temporal `\textit{until}' operator. Then, an independent judge model evaluates the candidate specification against strict, quantitative rubrics. The judge assigns discrete scores evaluating the inclusion of all necessary subjects and events, the correct temporal alignment of these entities as implied by the reference answer, and the syntactic validity of the chosen logical operators.

            \begin{table}[ht]
            \centering
            \caption{Training and rollout hyperparameters. Common settings apply to both \texttt{VSeek} unless noted.}
            \label{tab:exp_hparams}
            \small
            \begin{tabular}{p{0.40\linewidth}p{0.18\linewidth}p{0.18\linewidth}p{0.18\linewidth}}
                \toprule
                
                \textbf{Hyperparameter} & \textbf{GRPO} & \textbf{RLOO} & \textbf{GSPO} \\
                \midrule
                Training steps & 900 & 900 & 900 \\
                LLM learning rate & 1e-6 & 1e-6 & 1e-6 \\
                Warm-up ratio (policy) & 0.285 & 0.285 & 0.285 \\
                Samples per prompt (rollouts) & 5 & 5 & 5 \\
                KL regularization coefficient $\beta$ & 0.001 & 0.001 & 0.001 \\
                Entropy Coefficient $\lambda$ & 0.0001 & 0.0001 & 0.0001 \\
                Clip ratio Low $\epsilon_{\text{low}}$ & 0.2 & 0.2 & 3e-4 \\
                Clip ratio High $\epsilon_{\text{high}}$ & 0.28 & 0.28 & 4e-4 \\
                Negative Advantage Clip $\epsilon_{\text{neg}}$ & 10.0 & 10.0 & 10.0 \\
                Train and Validation batch size & 32 & 32 & 32 \\
                Mini batch size & 32 & 32 & 32 \\
                PPO Micro batch size & 8 & 8 & 8 \\
                Max input sequence length (tokens) & 2536 & 2536 & 2536 \\
                Max response length (tokens) & 5556 & 5556 & 5556 \\
                Max Agent Steps $B$ (search steps) & 4 & 4 & 4 \\
                Retriever top-$k$ (Video Clips) & 4 & 4 & 4 \\
                Rollout temperature / top-$p$ & 1.0 / 1.0 & 1.0 / 1.0 & 1.0 / 1.0 \\
                vLLM tensor-parallel / mem util. & 1 / 0.6 & 1 / 0.6 & 1 / 0.6 \\
                Checkpoint save frequency (steps) & 40 & 40 & 40 \\
                Evaluation Mode & \multicolumn{3}{l}{Evaluate most recent stable checkpoint} \\
                Compute and parallelism & \multicolumn{3}{l}{Single node, 8$\times$GH200; FSDP with CPU offload} \\
                
                \bottomrule
            \end{tabular}
        \end{table}
        
        If the initial specification exhibits deficiencies, the judge generates structured, diagnostic feedback along with a suggested revision. To balance computational cost with specification quality, we constrain this refinement cycle to exactly two iterations. During the subsequent feedback-driven pass, a reasoning module consumes the original question-answer context, the first-pass specification output, and the judge's diagnostic feedback to synthesize a refined, logically coherent final specification. This step ensures the resulting formulations accurately ground the subsequent reinforcement learning rewards without introducing hallucinated visual targets or logically contradictory sequence triggers.

        Finally, during the dense reward computation phase (Phase 2), the extracted logical primitives from the accepted specifications are matched against the visual observations. This matching process relies on evaluating the semantic similarity between the text-based primitives and the dense visual embeddings of the video clips, which are actively obtained using our base retriever, the ViCLIP model. This reward computation happens during the tool call process. To comprehensively evaluate the agent's video search behavior, the semantic matching accuracies for each primitive are systematically compiled and aggregated across all video search and tool call steps within the trajectory. This aggregated propositional accuracy directly informs the \ac{VETL} reward, effectively incentivizing the model to successfully surface the necessary visual components across its complete search sequence.
    
    \subsection{Training Hyperparameters} 
    
        We outline the training hyperparameters for each variant of \texttt{VSeek} in terms of the \ac{RL} post-training algorithms used in Table \ref{tab:exp_hparams}.

    \subsection{Prompts}

    This section details the prompts used to obtain the temporal logic decomposition of the question and its answer, as well as the prompts used during evaluation and \ac{RL} training.
    
    \subsubsection{VETL Prompts}

    \paragraph{Initial Temporal Logic Generator.} This prompt is used to generate the first specification when no judge feedback is present.
    
    \begin{tcolorbox}[breakable,colback=gray!5!white,colframe=black!75!black,title=Initial Temporal Logic Generator Prompt,fonttitle=\bfseries]
    \scriptsize
    \begin{Verbatim}[breaklines,breakanywhere]
    You are an intelligent agent that extracts atomic primitives 
    from a video QA sample and generates a 
    temporal-logic specification.
    
    IMPORTANT:
    
    The input question may already contain options. Use "Correct Answer"
    text as the source of truth whenever it is provided. If only an 
    answer index is provided, use it to map to the option in the question.
    
    Stage 1: Proposition Extraction
    
    Extract atomic primitives from: (a) the question context 
    and (b) the correct answer text.
    Each proposition must describe exactly one atomic event/state.
    Never merge multiple events in one proposition.
    Do not hallucinate facts not in the question or answer text.
    Be concrete and specific (avoid vague text like "something happens").
    Keep proposition text concise and directly usable in logic.
    If subtitle/caption content is referenced, use: 
    subtitle_'<exact text>'.
    You must only use the correct answer text as the source of 
    truth for the primitives. Do not use distractor options.
    
    Stage 2: Specification Generation
    
    Generate one specification using ONLY primitives from Stage 1.
    
    Hard rules:
    
    Use each proposition exactly once. 
    Use ONLY operators: AND, OR, NOT, UNTIL.
    Do not introduce new primitives.
    The specification must reflect the question's temporal intent.
    
    Temporal-operator decision guide:
    
    Use UNTIL only when the question expresses a true sequence trigger:
    
    "before X" → answer\_event UNTIL X;
    "after X" / "what happens when X" → X UNTIL answer\_event;
    "between X and Y" → X UNTIL (answer\_event AND Y)
    
    Use AND for co-existing facts/attributes.
    Use OR only for explicit disjunction in the question/answer.
    Use NOT only for explicit negation in the question/answer.
    
    If no temporal cue exists, avoid UNTIL.
    
    Return ONLY valid JSON:
    
    \{
    "proposition": ["<proposition1>", "<proposition2>", "..."],
    "specification": "<specification\_in\_temporal\_logic>"
    \}
    \end{Verbatim}
    
    \end{tcolorbox}

    \paragraph{Temporal Logic Feedback-Based Generator.} 
    This prompt is used to regenerate the temporal logic specification based on the feedback from a judge model that evaluates the specification from a prior generation pass.
    
    \begin{tcolorbox}[breakable,colback=gray!5!white,colframe=black!75!black,title=Temporal Logic Feedback Based Generator,fonttitle=\bfseries]
    
    \scriptsize
    \begin{Verbatim}[breaklines,breakanywhere]
    You are improving a temporal-logic extraction result.
    
    Task:
    Use the original QA, the first-pass output, the judge's 
    feedback and the judge's proposed revision.
    Re-reason and produce the BEST final proposition/specification.
    Prefer judge feedback, but do not blindly copy if inconsistent.
    
    QA:
    question: question
    options: options
    correct choice: correct\_choice
    
    Previous-pass Output:
    proposition: proposition
    specification: specification
    
    
    Judge feedback:
    inclusion: inclusion
    temporal_alignment: temporal_alignment
    operator_correctness: operator_correctness
    verdict: verdict
    needs_revision: needs_revision
    judge_revised_proposition: judge_revised_proposition
    judge_revised_specification: judge_revised_specification
    
    Requirements:
    
    Return only JSON with keys "proposition" and "specification".
    Proposition must be a non-empty list of atomic strings.
    The specification must be non-empty and only use these 
    operators conceptually: AND, OR, NOT, UNTIL (symbols \& | ! U).
    Keep it grounded in the question and correct answer, 
    and avoid distractors.
    
    \end{Verbatim}
    
    \end{tcolorbox}

    \paragraph{Judge Verification.}
    
    The judge evaluates the primitive based on the event and temporal criterion outlined in the prompt. The judge also provides feedback to improve the temporal logic specification of the prompt and proposes changes to the specification based on its internal feedback.
    
    \begin{tcolorbox}[breakable,colback=gray!5!white,colframe=black!75!black,title=Prompt 2: Judge Evaluation and Feedback, fonttitle=\bfseries]
    \scriptsize
    \begin{Verbatim}[breaklines,breakanywhere]
    You are an expert evaluator of temporal logic extraction from video 
    QA. Evaluate whether this output is correct for the QA sample.
    
    QA Sample
    Question: {question}
    Options: {options_block}
    Correct Answer Index: {correct_choice}
    
    Predicted Output
    proposition: {proposition}
    specification: {specification}
    
    Evaluation criteria (score each from 0 to 2)
    Inclusion of all potential subjects and events
    2: Includes all key subjects/events needed; no critical missing item.
    Does not include distractor options.
    1: Mostly complete but missing at least one non-trivial subject/event.
    0: Misses multiple key subjects/events or is largely irrelevant.
    
    Temporal alignment of subjects/events
    2: The ordering/temporal relationship matches the QA.
    1: Partially aligned; one notable temporal mismatch or ambiguity.
    0: Temporal relationships are wrong or unsupported.
    
    Correct temporal operators between subjects and events
    2: Operators (AND/OR/NOT/UNTIL) are appropriate and consistent.
    1: Minor operator issue but mostly acceptable.
    0: Operator usage is incorrect, contradictory, or invalid.
    
    Important rules:
    
    Judge only against the QA sample and correct answer.
    Penalize hallucinated events that are not grounded in the 
    question/answer.
    Prefer atomic primitives over merged multi-event clauses.
    If the current output is wrong, provide a revised 
    proposition/specification.
    
    Return JSON with EXACTLY this schema:
    {
    "inclusion": {"score": 0, "reason": "", "missing_or_extra": []},
    "temporal_alignment": {"score": 0, "reason": ""},
    "operator_correctness": {"score": 0, "reason": ""},
    "overall_score": 0,
    "verdict": "correct",
    "needs_revision": false,
    "revised_proposition": [],
    "revised_specification": ""
    }
    \end{Verbatim}
    \end{tcolorbox}
    
    \subsubsection{VSeek Prompt}
    
    We outline the prompt that is used as the system prompt for \texttt{Qwen3-VL-4B-Thinking} to perform the interleaved search and question-answering task. We use the same prompt for the \texttt{VSeek-GPT} variant to enable \texttt{GPT-5.2} to utilize the search-based indexing tool framework.
    
    \begin{tcolorbox}[breakable,colback=gray!5!white,colframe=black!75!black,title=\texttt{VSeek} System Prompt, fonttitle=\bfseries]
    \scriptsize
    \begin{Verbatim}[breaklines,breakanywhere]
    
    You are a video analysis agent that would aim to answer the user's question over multiple turns. 
    You will also have access to a tool-based retrieval system to retrieve more relevant frames of interest from a video. 
                           
     **INSTRUCTIONS**:
    Follow these instructions precisely on every turn.
             
        1) Reason: You must write your concise reasoning of 100-200 words within <think>...</think> that first summarizes and analyzes the frames to answer the question or to decide if you need to write a search query for more information.
        2) Decide: Based on your reasoning, decide if you have enough information in the frames obtained so far to either search for more information or answer the question.
        3) Act: At each turn you must choose ONE of the following actions:
            - If you need to search for more information using language based search to find objects or actions, output <search>...</search>
            - If you need to search for frames that contain a specific subtitle, output <search_subtitle>...</search_subtitle>.
            - If you need frames sampled uniformly from the whole video if the question requires it, output <search_summary></search_summary>.
            - If you decide you have enough information in the frames obtained so far to answer the question, output exactly one <answer>...</answer>.
    
    **Tool Call Specification**"
        - Strictly follow the format of <search><string></search> or <search_subtitle><string></search_subtitle> or <search_summary></search_summary> to invoke a tool call.
        - <search_summary> is used to search for frames sampled uniformly from the whole video, it does not need to be followed by a string query.
        - <string> is the search query in natural language
        - Emit EXACTLY ONE <search> or <search_subtitle> or <search_summary> per turn only when you need more information; no extra text outside the tags.
    
    **Answer Specification**
        - When you have enough information, output ONLY the option number inside <answer>...</answer>
        - Options are numbered 0 to N or A to Z as mentioned in the question. Do not include any words, just the number or letter corresponding to the option.
    
    **Behavioral Rules**
        - Each turn constitutes one <think> block followed by one action block (<search> OR <search_subtitle> OR <search_summary> OR <answer>) followed by a response from the tool.
        - Always think inside <think> before any action block (<search> OR <search_subtitle> OR <search_summary> OR <answer>) after getting information from the tool. If you see no information from the tool, you should search for more information.
        - You must only call the <search_summary> once since all the frames are sampled uniformly from the whole video.
        - You can call the <search> or <search_subtitle> or <search_summary> or <answer> but not both in the same turn. The choice of <search> or <search_subtitle> or <search_summary> or <answer> can be different across turns. See examples below.
        - If you answer the question, the trajectory ends.
        - You have a maximum 4 turns to answer the question.
    
    **Examples: Here are some correct examples of how to use the tool and answer the question**
    
     EXAMPLE 1 (Language search):
                
                Question: What is the chef doing? Answer options`: 0. Mixing flour 1. Pouring flour 2. Adding ingredients 3. Pouring flour
                <think>I should first locate where the chef uses a mixing bowl.</think>
                <search>a chef with a large mixing bowl</search>
                
                tool 
     (After the search tool call, the agent receives frames of the chef placing an empty bowl on the counter.)
                
                <think>I saw the bowl but no ingredient yet. I should fetch the next action of adding an ingredient.</think>
                <search>chef adding an ingredient to the bowl</search>
                
                tool 
     (After the second search, the agent receives frames of the chef pouring flour into the bowl.)
                <think>These frames show flour being added instead of being mixed. I don't see other ingredients. I can answer now.</think>
                <answer>3</answer>
                
    EXAMPLE 2 (Temporal reasoning with language search):
                
                Question: Who does the person throw the ball to? Answer options: 0. The cat 1. The dog 2. The person 3. The ball
                
                <think>I need to see the entire video first to answer the question.</think>
                <search_summary></search_summary>
    
                tool 
    (After the search_summary tool call, the agent receives frames uniformly sampled from the whole video.)
                <think>I see the person throwing the ball to the dog. I can answer now.</think>
                <answer>1</answer>
                
    EXAMPLE 3 (Subtitle-guided search):
                
                Question: What is the woman doing after the subtitle 'you're interested in.' appears? Answer options: A. A hat B. A scarf C. A jacket D. A dress
                
                <think>I should locate the moment the subtitle 'you're interested in.' appears.</think>
                <search_subtitle>you're interested in.</search_subtitle>
                
                tool 
     (After the first search, the agent receives frames corresponding to the subtitle)
                <think>The frames around the subtitle disambiguate. I will now search for woman clothing after the subtitle.</think>
                <search>a woman wearing a hat</search>
                
                tool 
     (After the second search, the agent receives frames of the woman wearing a hat.)
                <think>The woman is wearing a hat. I can answer now.</think>
                <answer>A</answer>
                
    \end{Verbatim}
    \end{tcolorbox}

    \subsubsection{Evaluation Prompts for Base Models}
    
    \paragraph{VideoQA Answering Prompts.}
    We use two closely related prompting templates depending on whether the underlying model is a base instruction-following \ac{LLM} or a reasoning-focused ``thinking'' variant.
    
    \begin{tcolorbox}[breakable,colback=gray!5!white,colframe=black!75!black,title=VideoQA Prompt (Base Instruct Models),fonttitle=\bfseries]
    \scriptsize
    \begin{Verbatim}[breaklines,breakanywhere]
    You are a helpful assistant. Look at the provided images sampled uniformly from a video and must choose the correct option from the given options to answer the question.
    You must only output the number of the correct option without thinking. Eg. 3
    \end{Verbatim}
    \end{tcolorbox}
    
    \begin{tcolorbox}[breakable,colback=gray!5!white,colframe=black!75!black,title=VideoQA Prompt (Thinking Models),fonttitle=\bfseries]
    \scriptsize
    \begin{Verbatim}[breaklines,breakanywhere]
    You are a helpful assistant. Look at the provided images sampled uniformly from a video and answer the question concisely.
    **INSTRUCTIONS**:
    1. First think very concisely about the question and the provided images within the <think> and </think> tags.
    2. You must provide the final answer the question with the correct option after ###.
    3. Do not provide empty fields and you must provide an answer to the best of your ability.
    4. For example.
    <think>I have found the mixing bowl, but no ingredients have been added yet. I need to find the next action where something is put into the bowl. I found it, based on the question, the action is mixing, hence the answer option is 3.</think>
    ### 3
    \end{Verbatim}
    \end{tcolorbox}



\section{Appendix: Qualitative Examples on Reasoning Traces}

\subsection{Reasoning Traces for Successes}

%

\begingroup
\scriptsize
\setlength{\tabcolsep}{6pt}
\renewcommand{\arraystretch}{1.15}
\begin{longtable}{p{0.15\linewidth} p{0.77\linewidth}}
\caption{\texttt{VSeek} confirming the existence of an event to answer accurately}\label{tab:qual-trace-template1}\\
\toprule
\textbf{Turn / Field} & \textbf{Content} \\
\midrule
\endfirsthead

\toprule
\textbf{Turn / Field} & \textbf{Content} \\
\midrule
\endhead

\midrule
\multicolumn{2}{r}{\emph{Continued on next page}}\\
\endfoot

\bottomrule
\endlastfoot

Question & On a sea with white waves rolling, a man slightly bends his knees, leans back, and stands on a light blue surfboard to surf. After this, what happens next?\par\smallskip
\noindent 0. A person in a blue swimsuit jumps into the water
\par\smallskip
\noindent 1. A person in a black swimsuit jumps into the water
\par\smallskip
\noindent 2. Two fully armed people walk on a forest path
\par\smallskip
\noindent 3. A person in a swimsuit walks on the beach
\\

\\
Original Video  &
\includegraphics[width=1.0\linewidth]{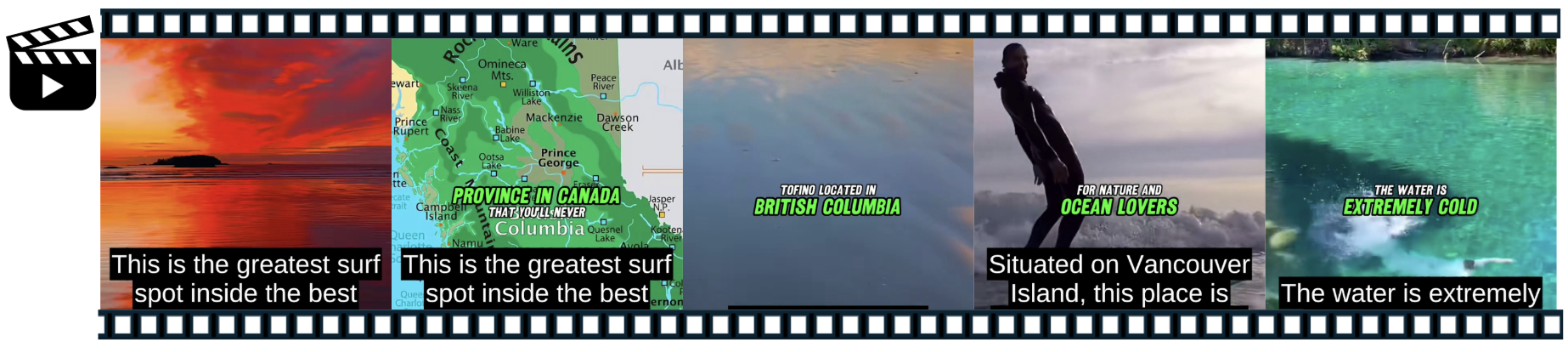}

\\
Turn 1 &
\textbf{Reasoning/output:} \texttt{<think>} Okay, let's try to figure out this problem. The question is about what happens next after a man is surfing on a light blue surfboard with white waves. The options are 0, 1, 2, 3. First, let me look at the current scenario. The man is surfing, which is on the sea. The next action should be related to someone entering the water or moving on the beach. Option 0: A person in a blue swimsuit jumps into the water. Option 1: Black swimsuit. Option 2: Two armed people on a forest path. Option 3: A person walking on the beach. So the current situation is the surfer, and the next step could be someone jumping in. But which swimsuit? Or maybe the surfer is about to do something else. Alternatively, maybe the person is already on the board, and the next frame shows someone else entering the water. Alternatively, perhaps the correct answer is 0 or 1. Let me check the options again. If the man is surfing, then the next action might be another person jumping in, so either 0 or 1. But which one? Or perhaps it's a different action, like someone walking on the beach (option 3). But the surfer is already on the water, so maybe the next step is someone entering the water. Alternatively, perhaps the correct answer is 0 or 1, so the answer is either 0 or 1. But which one? Let's see if there's a way to determine which swimsuit. Alternatively, maybe the answer is 0, as the blue swimsuit is common. Alternatively, perhaps the correct answer is 0 or 1, but the options are given as 0 and 1, so the correct choice would be one of them. Alternatively, maybe the next action is someone walking on the beach (option 3), but the surfer is already in the water, so perhaps the person is on the beach. Hmm, not sure. Let me think again. Alternatively, perhaps the correct answer is 0 or 1, so the answer is 0 or 1. \texttt{</think> <search>} a person jumping into the water \texttt{</search>} \\

& \textbf{Video Search Tool Response}\\
&\includegraphics[width=0.8\linewidth]{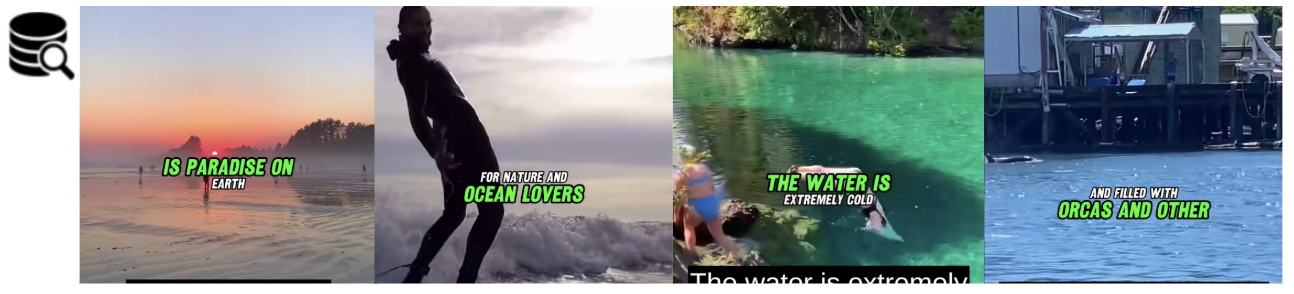}

\\
Turn 2

& \textbf{Reasoning/output:} \texttt{<think>}Looking at the frames, there's a person in a black swimsuit (option 1) jumping into the water, which aligns with option 1. Alternatively, maybe the person is wearing a black swimsuit. So the correct answer would be 1. \texttt{</think>} \texttt{<search>}person in black swimsuit \texttt{</search>} \\

&\textbf{Video Search Tool Response:}\\
& \includegraphics[width=0.8\linewidth]{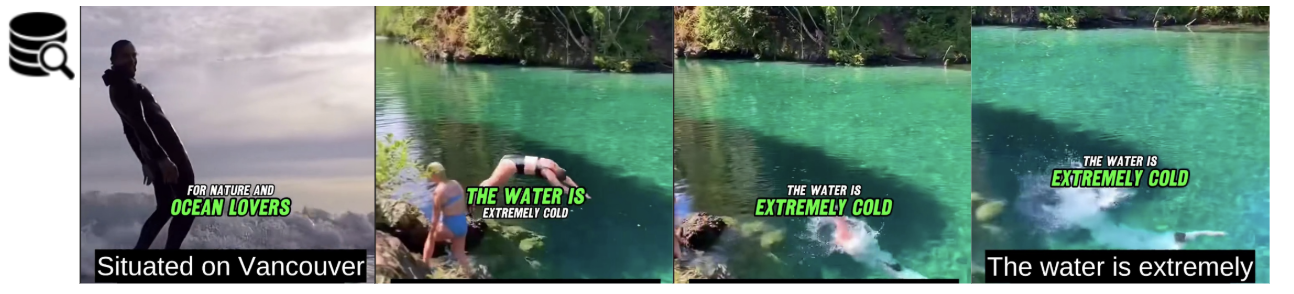}

\\
Turn 3  
&
\textbf{Reasoning/output:}
\texttt{<think>} Now that we've confirmed the person is in a black swimsuit, which matches option 1, the correct answer is 1. \texttt{</think>} \texttt{<answer>}1\texttt{</answer>}\\

\textbf{Final answer:} & Option 1\\

\textbf{Ground Truth} & Option 1 
\end{longtable}
\endgroup

\begingroup
\scriptsize
\setlength{\tabcolsep}{6pt}
\renewcommand{\arraystretch}{1.15}
\begin{longtable}{p{0.15\linewidth} p{0.77\linewidth}}
\caption{\texttt{VSeek} confirming the existence of an object to answer accurately}\label{tab:qual-trace-template}\\
\toprule
\textbf{Turn / Field} & \textbf{Content} \\
\midrule
\endfirsthead

\toprule
\textbf{Turn / Field} & \textbf{Content} \\
\midrule
\endhead

\midrule
\multicolumn{2}{r}{\emph{Continued on next page}}\\
\endfoot

\bottomrule
\endlastfoot

Question & In the kitchen, after a chef pours a yellow liquid into a wok with stir-fried vegetables, what happens on the screen? 
\par\smallskip
\noindent 0. The chef adds eggplant to the wok
\par\smallskip
\noindent 1. The chef transfers the vegetables from the wok into a rectangular white plate containing bread.
\par\smallskip
\noindent 2. The chef adds scallions to the wok.
\par\smallskip
\noindent 3. The chef transfers the vegetables from the wok into a rectangular white plate containing rice.
\par \smallskip
\noindent 4. The chef transfers the vegetables from the wok into a round white plate containing rice.
\\

\\
Original Video  &
\includegraphics[width=1.0\linewidth]{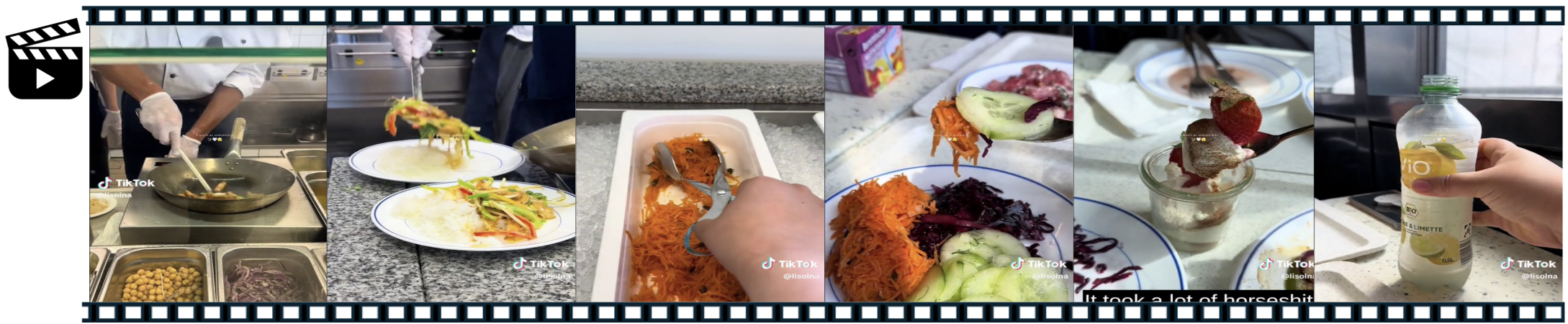}

\\
Turn 1 &
\textbf{Reasoning/output:} \texttt{<think>} Okay, let's try to figure out this problem. The question is about what happens after a chef pours a yellow liquid into a wok with stir-fried vegetables. The options are 0 to 4, each describing different actions. First, I need to determine which option correctly shows the next step. Let me look at the options again: 0) Add eggplant 1) Transfer vegetables to a rectangular white plate with bread 2) Add scallions 3) Transfer to rectangular white plate with rice 4) Transfer to round white plate with rice So after pouring the yellow liquid (maybe something like sauce or egg), the next action is transferring the vegetables. The key is the plate's shape and contents. If the correct option is 3 or 4, which one matches the plate. Let's think: if the plate is rectangular and has rice, then 3. If round and rice, then 4. Alternatively, maybe the plate is rectangular with rice, so 3. So the answer would be 3 or 4. Let me check the options again. So the correct answer is either 3 or 4. Based on the possible frames, the correct one is 3 or 4. So the answer should be 3 or 4. Lets summarize. Reason: After pouring the yellow liquid, the next step is transferring the vegetables to a plate. The correct option depends on the plate's shape and contents. Options 3 and 4 both involve rice, but the plate's shape (rectangular vs. round) is critical. Decide: Enough information is available to choose between 3 and 4. Need to confirm which one is correct. Act: Use video search to find the specific frame where the transfer happens and check the plate's details. \texttt{</think> <search>} chef transferring vegetables to a plate \texttt{</search>} \\

& \textbf{Video sample:}\\
&\includegraphics[width=0.8\linewidth]{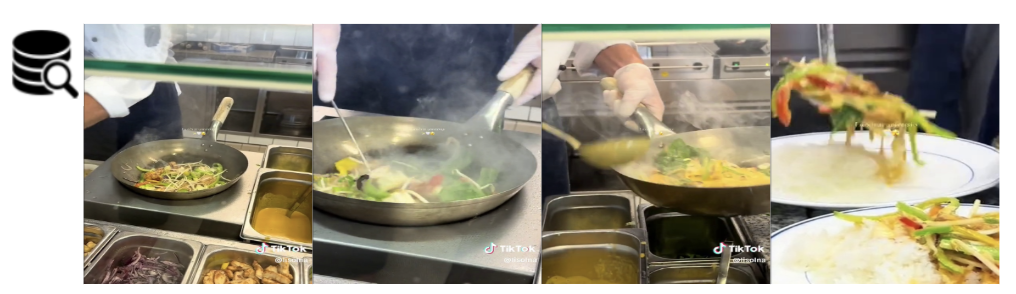}

\\
Turn 2

& \textbf{Reasoning/output:} \texttt{<think>}  Looking at the frames, the vegetables are being transferred to a plate, and the plate appears to be round (option 4) and contains rice. Alternatively, maybe it's a rectangular plate, but the frames show a round plate. So option 4: transferring to a round white plate with rice. \texttt{</think>}  \texttt{<search>}  round white plate containing rice \texttt{</search>}  \\

&\textbf{Video sample:}\\
&\includegraphics[width=0.8\linewidth]{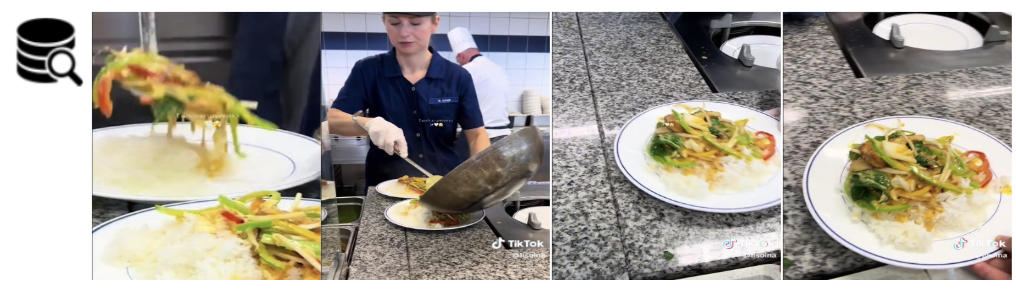}

\\
Turn 3  &
\textbf{Video sample:}~

\textbf{Reasoning/output:}
\texttt{<think>}  Now it's clear that the plate is round and contains rice, matching option 4. \texttt{</think>}  \texttt{<answer>} 4 \texttt{</answer>} \\

\textbf{Final answer:} & Option 4\\
\textbf{Ground Truth} & Option 4 \\
\end{longtable}
\endgroup

\begingroup
\scriptsize
\setlength{\tabcolsep}{6pt}
\renewcommand{\arraystretch}{1.15}
\begin{longtable}{p{0.15\linewidth} p{0.77\linewidth}}
\caption{\texttt{VSeek} confirming the existence of an event to answer accurately}\label{tab:qual-trace-example1}\\
\toprule
\textbf{Turn / Field} & \textbf{Content} \\
\midrule
\endfirsthead

\toprule
\textbf{Turn / Field} & \textbf{Content} \\
\midrule
\endhead

\midrule
\multicolumn{2}{r}{\emph{Continued on next page}}\\
\endfoot

\bottomrule
\endlastfoot

Question & Inside a room, there is a bookshelf filled with books and a wall covered in wallpaper. A man wearing a red short sleeve shirt is sitting on a gaming chair. When mentioning 'smd, what is caught? Piper, aka resource, is a trace format multi,' what is this man doing?
\par\smallskip
\noindent 0. He has both hands raised above his head
\par\smallskip
\noindent 1. He has both hands crossed in front of his chest
\par\smallskip
\noindent 2. He is doing the V-sign with both hands
\par\smallskip
\noindent 3. He is doing the V-sign with one hand and clenching a fist with the other
\\

\\
Original Video &
\includegraphics[width=1.0\linewidth]{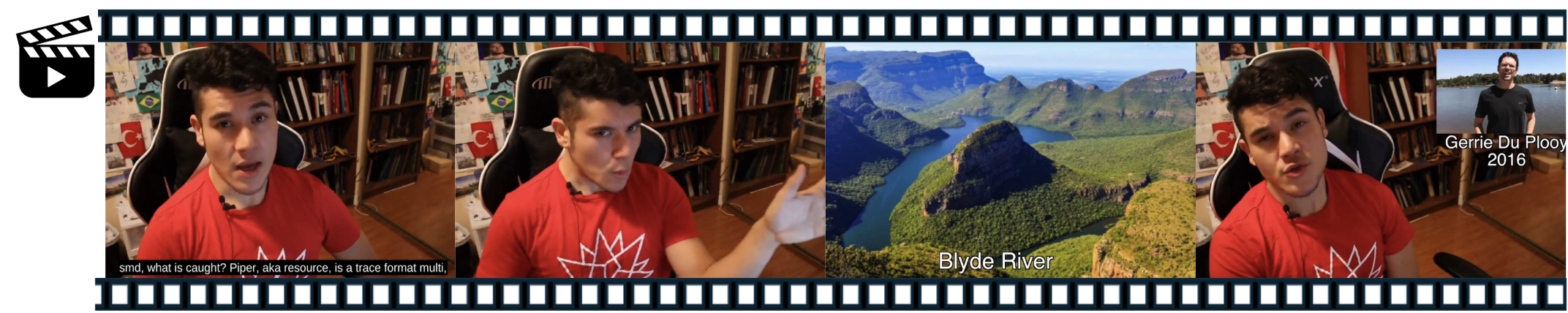}

\\
Turn 1 &
\textbf{Reasoning/output:} \texttt{<think>} Okay, let's try to figure out this problem. So the question is about a man in a room with a bookshelf, wallpaper, and he's wearing a red short-sleeve shirt sitting on a gaming chair. The subtitle mentions "smd, what is caught? Piper, aka resource, is a trace format multi," and we need to determine what he's doing based on the options 0 to 3. First, let's look at the options: 0) Both hands raised above head 1) Both hands crossed in front of chest 2) V-sign with both hands 3) V-sign with one hand and clenching a fist with the other The key here is to see which of these gestures aligns with the subtitle. The subtitle is a bit cryptic, but maybe the action he's making is a specific sign. Let's think about the possible gestures. If he's doing the V-sign (which is two fingers up, like "V"), then that would be option 2 or 3. Option 2 is V-sign with both hands, option 3 is V with one and a fist with the other. Alternatively, maybe the "smd" part refers to something else, but perhaps the main point is the gesture. Let's check the frames. If the frames show him making a V with both hands, then 2. If it's one V and one fist, then 3. Alternatively, perhaps he's making a different gesture. So the correct answer would be either 2 or 3. Let me check which one matches. \texttt{</think> <search>} a man doing a V-sign with both hands \texttt{</search>} \\

& \textbf{Video sample:} \\
& \includegraphics[width=0.8\linewidth]{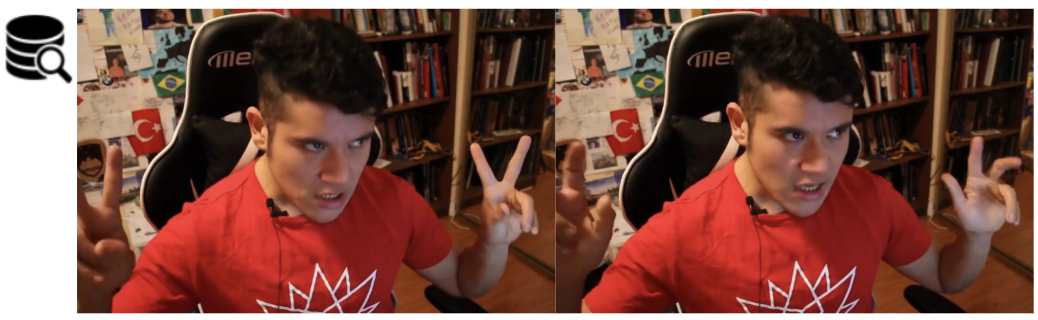}

\\
Turn 2 &
\textbf{Reasoning/output:} \texttt{<think>} Looking at the frames, the person is making a gesture with both hands. The frames show him raising his hands, which could be the V-sign with both hands (option 2). Alternatively, maybe he's doing a different gesture. Let me check again. Alternatively, maybe he's making a "V" with both hands, which would be option 2. So the answer is 2. \texttt{</think>} \texttt{<search>} both hands making V-sign \texttt{</search>} \\

&\textbf{Video sample:}\\
&\includegraphics[width=0.8\linewidth]{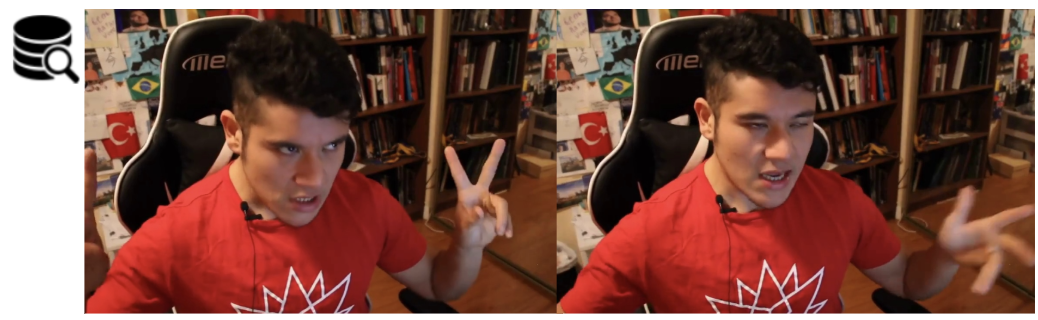}

\\
Turn 3 &
\textbf{Video sample:}~

\textbf{Reasoning/output:}
\texttt{<think>} Now it's clear that he is making the V-sign with both hands, which matches option 2. \texttt{</think>} \texttt{<answer>} 2 \texttt{</answer>} \\

\textbf{Final Answer} & Option 2\\
\textbf{Ground Truth} & Option 2 \\
\end{longtable}
\endgroup

\begingroup
\scriptsize
\setlength{\tabcolsep}{6pt}
\renewcommand{\arraystretch}{1.15}
\begin{longtable}{p{0.15\linewidth} p{0.77\linewidth}}
\caption{\texttt{VSeek} answering event-ordering questions from MLVU}\label{tab:qual-trace-example2}\\
\toprule
\textbf{Turn / Field} & \textbf{Content} \\
\midrule
\endfirsthead

\toprule
\textbf{Turn / Field} & \textbf{Content} \\
\midrule
\endhead

\midrule
\multicolumn{2}{r}{\emph{Continued on next page}}\\
\endfoot

\bottomrule
\endlastfoot

Question & Arrange the following events from the video in the correct chronological order: (1)Man is on lake side talking to the camera like other couples as well; (2)Man is talking to the camera; (3)People are kayaking on calm river and have a good picnic day; (4)People are standing on a side of a rock wall.
\par\smallskip
\noindent 0. 1→2→3→4
\par\smallskip
\noindent 1. 2→1→4→3
\par\smallskip
\noindent 2. 4→2→1→3
\par\smallskip
\noindent 3. 2→4→1→3

\\
Original Video &
\includegraphics[width=1.0\linewidth]{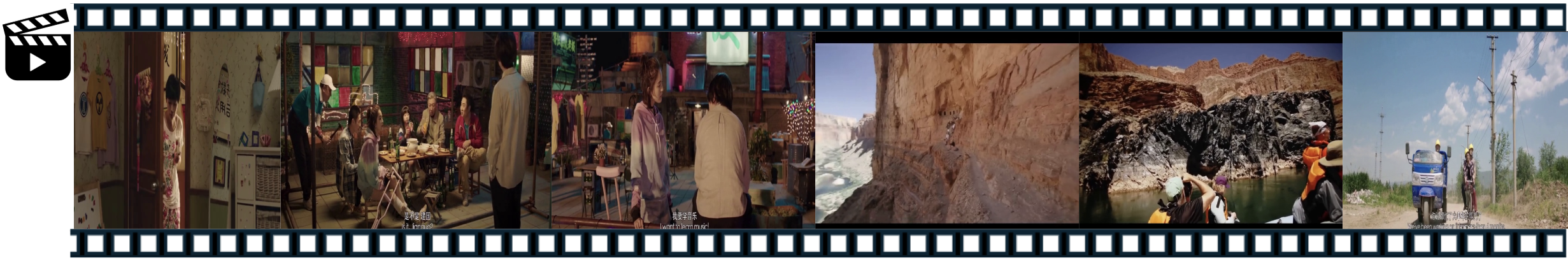}

\\
Turn 1 &
\textbf{Reasoning/output:} \texttt{<think>} Okay, let's try to figure out the correct chronological order of the events from the video. The options are given as 0, 1, 2, 3, each with different sequences. First, let's list the events again: 1. Man is on lake side talking to the camera like other couples as well. 2. Man is talking to the camera. 3. People are kayaking on calm river and have a good picnic day. 4. People are standing on a side of a rock wall. So, the task is to arrange these four events in the correct order. Looking at the options: Option 0: 1→2→3→4 Option 1: 2→1→4→3 Option 2: 4→2→1→3 Option 3: 2→4→1→3 Let me think. Which event comes first? If the man is talking to the camera (event 2), then perhaps he's the one on the lake side (event 1). Or maybe event 4 (people standing on rock wall) comes first. Alternatively, maybe the people are standing on the rock wall (4), then the man talks (2), then he's on the lake side (1), then the kayaking (3). So 4 → 2 →1 →3 would be option 2. Alternatively, if the man is talking (2), then he's on the lake side (1), then people are standing (4), then kayaking (3). But that would be 2→1→4→3 which is option 1. Hmm. Let's check the events again. Event 1: Man is on lake side talking to the camera like other couples as well. Event 2: Man is talking to the camera. So perhaps event 2 is the same as event 1, but maybe event 2 is the initial talking, then event 1 is part of the same scene. Or maybe event 2 is a different instance. Alternatively, perhaps the man is talking (2), then he's on the lake side (1), then the people are standing (4), then kayaking (3). So 2 →1 →4 →3, which is option 1. But another possibility: the people are standing on the rock (4), then the man talks (2), then he's on the lake (1), then kayaking (3). So 4 →2 →1 →3, which is option 2. Alternatively, maybe the man is talking (2), then the people are standing (4), then the man is on the lake (1), then kayaking (3). So 2 →4 →1 →3, which is option 3. So which one is correct? Alternatively, perhaps the sequence is: First, the people are standing on the rock wall (4), then the man talks (2), then he's on the lake (1), then they kayak (3). So 4,2,1,3 → option 2. Alternatively, maybe the man is talking (2), which is part of the lake side (1), so 2 is a subset of 1, but perhaps event 1 is the man on the lake side, which includes talking. Alternatively, perhaps the correct order is 2 (man talking), then 1 (man on lake side, which is part of the same talking), then 4 (people standing), then 3 (kayaking). But that would be 2→1→4→3, which is option 1. Alternatively, maybe the people are standing (4) first, then the man talks (2), then he's on the lake (1), then kayaking (3). So 4,2,1,3 → option 2. So which is correct? Let me think again. If the man is talking (2), then he's on the lake side (1), which is part of the same scene. So perhaps 2 comes first, then 1. Then the people are standing (4), then kayaking (3). So 2→1→4→3, which is option 1. Alternatively, perhaps the people standing (4) is before the man talks (2), so 4→2→1→3, which is option 2. Hmm. Alternatively, maybe the man is talking (2), then the people are standing (4), then he's on the lake (1), then kayaking (3). So 2→4→1→3, which is option 3. Alternatively, perhaps the correct order is 2, then 4, then 1, then 3. So option 3. But I need to check which one makes sense. Alternatively, perhaps the correct answer is 2→4→1→3 (option 3), but not sure. Alternatively, maybe the man is first talking (2), then the people are standing (4), then he's on the lake (1), then they kayak (3). So 2,4,1,3. So option 3 is 2→4→1→3, which would be correct. Alternatively, perhaps the man is on the lake (1) first, then talking (2), but that seems redundant. Alternatively, maybe the correct order is 4 (people standing), then 2 (man talking), then 1 (man on lake), then 3 (kayaking). So 4→2→1→3, which is option 2. So which is it? Let me think of the possible sequences. If the people are standing on the rock wall (4), then the man talks (2), then he's on the lake (1), then they kayak (3). So 4,2,1,3. So option 2. Alternatively, maybe the man is talking (2), then he's on the lake (1), which is part of the same scene. So 2→1, then 4 (people standing), then 3 (kayaking). So 2→1→4→3, which is option 1. So which is correct? Alternatively, perhaps the sequence is 2 (man talking), then 4 (people standing), then 1 (man on lake), then 3 (kayaking). So 2→4→1→3, option 3. Hmm. Maybe the correct answer is 3. Reason: Let's analyze the events. Event 2 is the man talking to the camera, which could be the initial action. Then, maybe the people are standing on the rock wall (event 4), then the man is on the lake side (event 1), and finally, they kayak (event 3). So the order would be 2 →4 →1 →3, which matches option 3. Decide: Enough information is obtained to determine the correct sequence. Act: \texttt{</think> <search>} man talking to the camera then people standing on rock wall \texttt{</search>} \\

& \textbf{Video sample:}\\
&\includegraphics[width=0.8\linewidth]{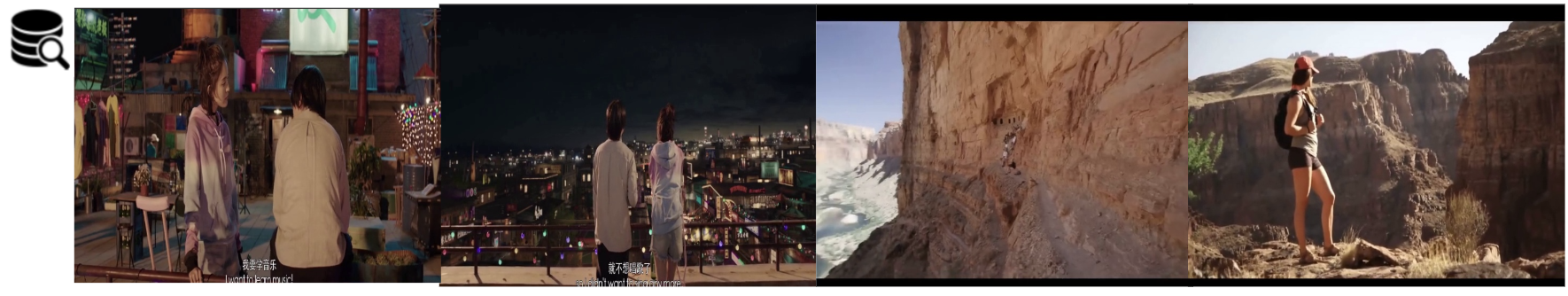}

\\
Turn 2 &
\textbf{Reasoning/output:} \texttt{<think>} From the frames, the first frame shows people standing on the rock wall (event 4), then the man talks (event 2), followed by the man on the lake side (event 1), then the kayaking (event 3). So the order is 4 →2 →1 →3, which is option 2. But wait, the frames also show people kayaking and having a picnic. Alternatively, perhaps the man is talking (2), then people standing (4), then he's on the lake (1), then kayaking (3). Hmm, maybe the correct order is 2→4→1→3 (option 3). Let me check again. Alternatively, the frames show the man talking (2), then people standing (4), then the lake side (1), then kayaking (3). So yes, 2→4→1→3, which is option 3. \texttt{</think>} \texttt{<search>} man talking then people standing on rock wall then lake side \texttt{</search>} \\

&\textbf{Video sample:}\\
&\includegraphics[width=0.8\linewidth]{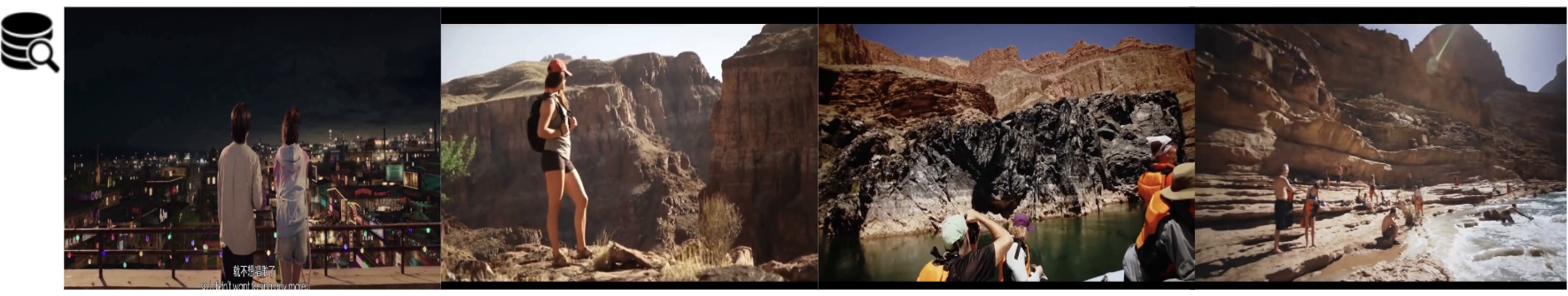}

\\
Turn 3 &
\textbf{Video sample:}~

\textbf{Reasoning/output:}
\texttt{<think>} Now, the frames confirm that after the man talks (2), people are standing (4), then the man is on the lake (1), then kayaking (3). So the order is 2 →4 →1 →3, which is option 3. \texttt{</think>} \texttt{<answer>} 3 \texttt{</answer>} \\
\textbf{Final Answer} & Option 3\\
\textbf{Ground Truth} & Option 3 \\

\end{longtable}
\endgroup

\begingroup
\scriptsize
\setlength{\tabcolsep}{6pt}
\renewcommand{\arraystretch}{1.15}
\begin{longtable}{p{0.15\linewidth} p{0.77\linewidth}}
\caption{\texttt{VSeek} answering event-ordering questions from MLVU}\label{tab:qual-trace-example3}\\
\toprule
\textbf{Turn / Field} & \textbf{Content} \\
\midrule
\endfirsthead

\toprule
\textbf{Turn / Field} & \textbf{Content} \\
\midrule
\endhead

\midrule
\multicolumn{2}{r}{\emph{Continued on next page}}\\
\endfoot

\bottomrule
\endlastfoot

Question & Which of the following options correctly matches the sequence of actions as they actually appear in the video?
\par\smallskip
\noindent 0. cooking sausages → baking cookies → milking cow → making jewelry
\par\smallskip
\noindent 1. making jewelry → cooking sausages → milking cow → baking cookies
\par\smallskip
\noindent 2. cooking sausages → making jewelry → baking cookies → milking cow
\par\smallskip
\noindent 3. baking cookies → making jewelry → milking cow → cooking sausages
\\
Original Video &
\includegraphics[width=1.0\linewidth,trim=2mm 0mm 0mm 2mm,clip]{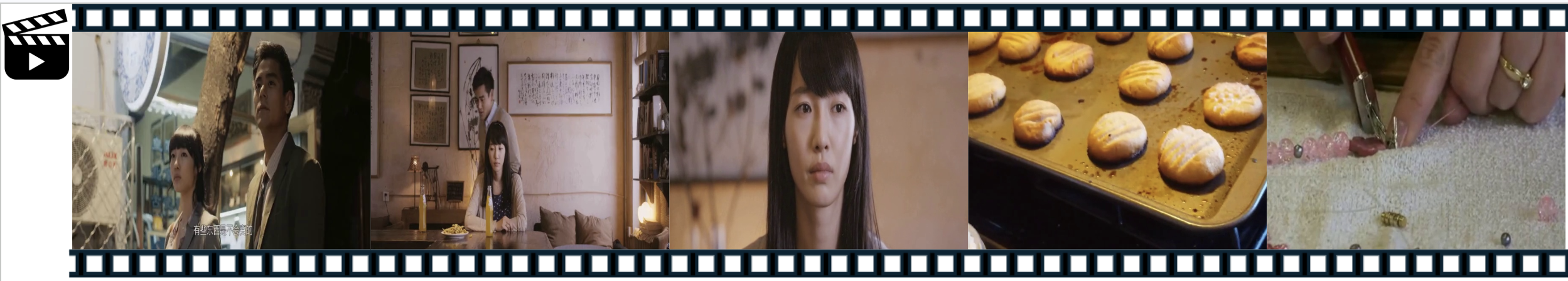}

\\
Turn 1 &
\textbf{Reasoning/output:} \texttt{<think>} Okay, let's try to figure out this problem. The user is asking which option correctly matches the sequence of actions from the video. The options are 0, 1, 2, 3. First, I need to determine the correct order of the four actions: cooking sausages, baking cookies, milking cow, making jewelry. Let me check each option: Option 0: cooking sausages $\rightarrow$ baking cookies $\rightarrow$ milking cow $\rightarrow$ making jewelry. So after cookies, then milking, then jewelry. Option 1: making jewelry $\rightarrow$ cooking sausages $\rightarrow$ milking cow $\rightarrow$ baking cookies. Option 2: cooking sausages $\rightarrow$ making jewelry $\rightarrow$ baking cookies $\rightarrow$ milking cow. Option 3: baking cookies $\rightarrow$ making jewelry $\rightarrow$ milking cow $\rightarrow$ cooking sausages. So which order is correct? Without seeing the video, but maybe I can use the video\_search function to find the correct frames. So I should call video\_search with a query that's a bit general, like "sequence of actions", and mode as base. That way, it can find the frames where the correct order is shown. \texttt{</think> <search>} a sequence of actions: cooking sausages, baking cookies, milking cow, making jewelry \texttt{</search>} \\

& \textbf{Video sample:}\\
&\includegraphics[width=0.8\linewidth]{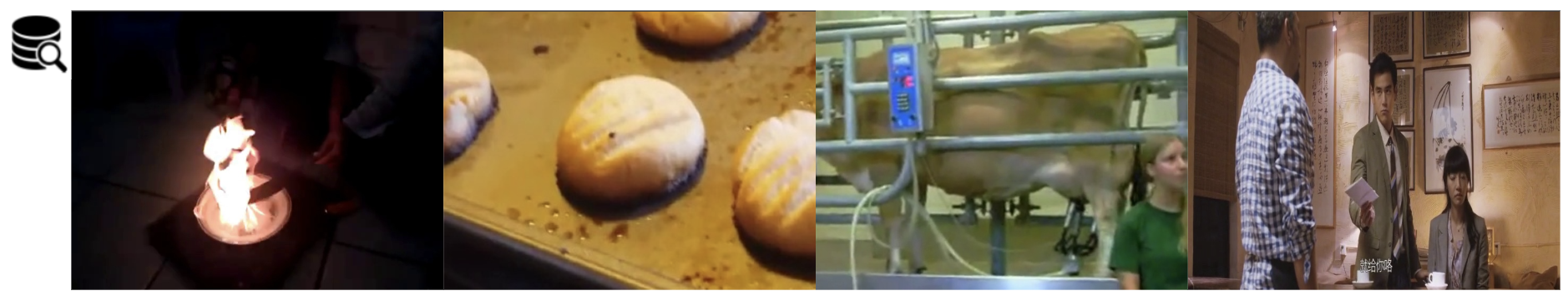}

\\
Turn 2 &
\textbf{Reasoning/output:} \texttt{<think>} From the frames, I can see that first there are cookies being baked (so baking cookies comes early), then milking the cow, and then making jewelry. Let's check the order: Baking cookies (frames 2-5), then milking cow (frames 7-11), then making jewelry (frames 13-15). So the sequence is baking cookies $\rightarrow$ milking cow $\rightarrow$ making jewelry. But the options have the four steps. Let me check the options again. Option 0: cooking sausages $\rightarrow$ baking cookies $\rightarrow$ milking cow $\rightarrow$ making jewelry. But in the frames, there's no sausages. So perhaps the first step is baking cookies, then milking, then making jewelry. So maybe the correct option is 3: baking cookies $\rightarrow$ making jewelry $\rightarrow$ milking cow $\rightarrow$ cooking sausages? Wait, no. Alternatively, perhaps the correct order is baking cookies first, then milking, then making jewelry, but the options don't have that. Let me check the options again. Looking at the options: 0: cooking sausages $\rightarrow$ baking cookies $\rightarrow$ milking cow $\rightarrow$ making jewelry 1: making jewelry $\rightarrow$ cooking sausages $\rightarrow$ milking cow $\rightarrow$ baking cookies 2: cooking sausages $\rightarrow$ making jewelry $\rightarrow$ baking cookies $\rightarrow$ milking cow 3: baking cookies $\rightarrow$ making jewelry $\rightarrow$ milking cow $\rightarrow$ cooking sausages Wait, but the frames show baking cookies first (so step 1 is baking), then milking (step 2), then making jewelry (step 3). So the order would be baking cookies, then milking, then making jewelry. So which option fits? Let's see: Option 0: cooking sausages first, then baking, then milking, then making. But there's no sausages in the frames. Option 3: baking, making, milking, then cooking. But the frames show milking after baking and before making? Or maybe not. Alternatively, perhaps the correct order is baking, then milking, then making. So perhaps the correct option is 3, but I'm not sure. Alternatively, maybe the sequence is baking, then milking, then making. So which option has that? Let me check again. Alternatively, perhaps there's a step I'm missing. Let me look again at the frames. Frames 2-5: baking cookies. Frames 7-11: milking cow. Frames 13-15: making jewelry. So the order is baking $\rightarrow$ milking $\rightarrow$ making. So the correct sequence would be baking, then milking, then making. So which option has that? Let's check the options: Option 3: baking cookies $\rightarrow$ making jewelry $\rightarrow$ milking cow $\rightarrow$ cooking sausages. No, that's not matching. Option 0: cooking sausages $\rightarrow$ baking $\rightarrow$ milking $\rightarrow$ making. But there's no sausages. Option 2: cooking sausages $\rightarrow$ making jewelry $\rightarrow$ baking $\rightarrow$ milking. Not matching. Hmm, perhaps there's another step. Maybe the milking comes after baking, then making jewelry. So perhaps the correct option is 0, but there's a mistake. Alternatively, perhaps the correct option is 3, but not sure. Alternatively, perhaps the correct option is 3, but I need to check again. Alternatively, maybe the correct option is 0, but the sausages are part of the sequence. Alternatively, perhaps the correct answer is 0, but the frames don't show the sausages. So perhaps I need to search again for the exact sequence. \texttt{</think>} \texttt{<search>} milking cow and making jewelry \texttt{</search>} \\

&\textbf{Video sample:}\\
&\includegraphics[width=0.8\linewidth]{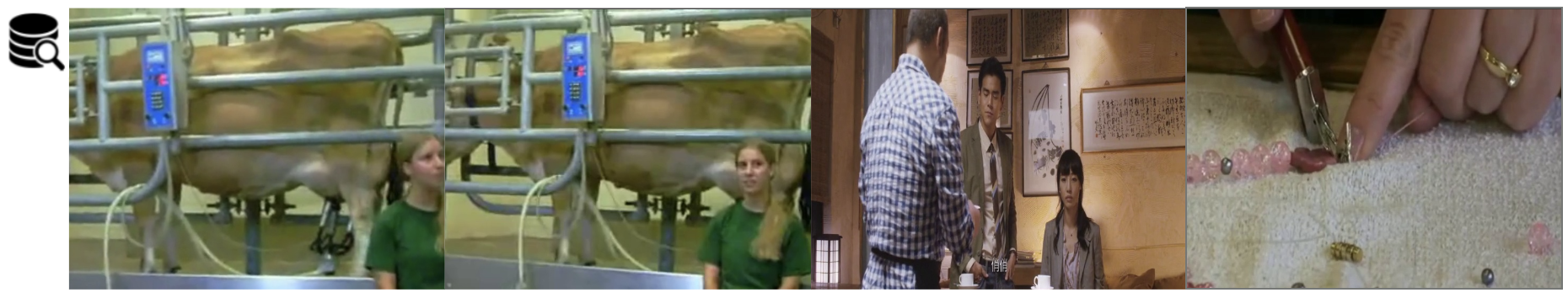}

\\
Turn 3 &
\textbf{Video sample:}~

\textbf{Reasoning/output:}
\texttt{<think>} Now, the frames confirm that after baking cookies (which was earlier), milking the cow comes next, then making jewelry. So the correct order is baking cookies $\rightarrow$ milking cow $\rightarrow$ making jewelry. Let's check the options again. Option 0: cooking sausages $\rightarrow$ baking $\rightarrow$ milking $\rightarrow$ making. But there's no sausages. Option 3: baking $\rightarrow$ making $\rightarrow$ milking $\rightarrow$ cooking. No. Wait, perhaps the correct option is 0, but the sausages are not present. Alternatively, maybe the correct option is 0, but the sausages are part of the sequence. Alternatively, since the other options don’t present baking $\rightarrow$  milking $\rightarrow$ making in this exact chronological sequence, perhaps the correct answer is 0, and the first step is cooking sausages, which isn't in the frames, but maybe it's part of the sequence. Hence, the answer is 0. \texttt{</think>} \texttt{<answer>} 0 \texttt{</answer>} \\

\textbf{Final Answer} & Option 0\\
\textbf{Ground Truth} & Option 0 \\
\end{longtable}
\endgroup

\begingroup
\scriptsize
\setlength{\tabcolsep}{4pt}
\renewcommand{\arraystretch}{1.1}
\begin{longtable}{p{0.15\linewidth} p{0.77\linewidth}}
\caption{\texttt{VSeek} searching correctly to answer an event-specific grounding question}\label{tab:qual-trace-example4}\\
\toprule
\textbf{Turn / Field} & \textbf{Content} \\
\midrule
\endfirsthead

\toprule
\textbf{Turn / Field} & \textbf{Content} \\
\midrule
\endhead

\midrule
\multicolumn{2}{r}{\emph{Continued on next page}}\\
\endfoot

\bottomrule
\endlastfoot

Question & In the scene, a woman is standing in front of a cash register, and there are two other people behind the counter. The four people in the scene are clearly visible. The woman at the register is wearing black clothes, and the woman buying coffee is wearing an olive-green trench coat. Who is the person in the scene with their head slightly bowed and smiling?
\par\smallskip
\noindent 0. The black-haired woman in the white jacket
\par\smallskip
\noindent 1. The black-haired woman in the olive-green jacket
\par\smallskip
\noindent 2. The man in the black shirt
\par\smallskip
\noindent 3. The woman in the olive-green trench coat
\par\smallskip
\noindent 4. The woman in the black dress
\\
Original Video &
\includegraphics[width=1.0\linewidth]{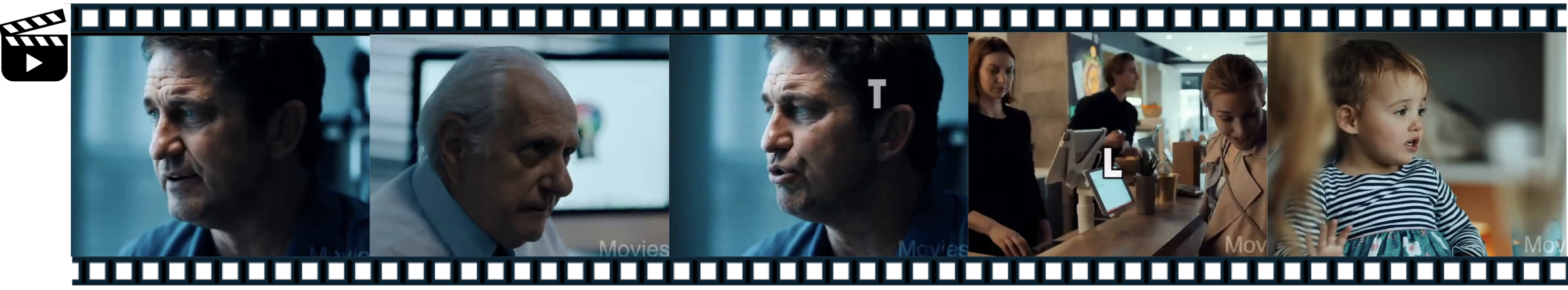}

\\
Turn 1 &
\textbf{Reasoning/output:} \texttt{<think>} Okay, let's try to figure out this problem. So, the question is asking which person in the scene has their head slightly bowed and is smiling. The options are 0 to 4. First, let's break down the given information. There's a woman at the cash register wearing black clothes, and the woman buying coffee is in an olive-green trench coat. The four people are visible, with two behind the counter. \\

& Looking at the options: 0) The black-haired woman in the white jacket 1) The black-haired woman in the olive-green jacket 2) The man in the black shirt 3) The woman in the olive-green trench coat (which is the buyer) 4) The woman in the black dress (the one at the register) The person with a head slightly bowed and smiling. Let's think. The woman at the register (option 4) is wearing black, but maybe she's the one. \\

& Alternatively, the buyer (option 3) could be smiling. But the description says the woman buying coffee is in the olive-green trench coat, which is option 3. Alternatively, maybe the woman in the black dress (option 4) is the one at the register, and she's the one with the bowed head. Wait, but the user says the woman at the register is wearing black clothes, so that's option 4. Alternatively, perhaps the black-haired woman in the olive-green jacket (option 1) is the one. Let me check again. \\

& Alternatively, maybe the person is the one in the white jacket (option 0), but the description says the woman at the register is in black, and the buyer is in olive-green. So the person with the bowed head and smile could be the one in the white jacket (option 0) or the buyer (option 3). Let me see which one makes sense. Alternatively, perhaps the woman in the olive-green trench coat (option 3) is the one smiling. So the answer would be 3. Or maybe the woman in the black dress (option 4) is the one at the register, so she's the one with the bowed head. Hmm. \\

& Let me check the options again. So, the answer options are 0,1,2,3,4. The correct choice is either 0,1,3, or 4. Let me think again. If the woman at the register is wearing black (option 4), and she's the one with the head bowed and smiling, then 4. Alternatively, the buyer (option 3) is the one. So which one is correct? Reason: The woman at the register is in black (option 4), and the buyer is in olive-green (option 3). The person with a bowed head and smile is likely the one at the register, so option 4. \\

& Alternatively, maybe the buyer is smiling. Let me confirm. Wait, perhaps the woman in the olive-green trench coat (option 3) is the buyer, and she's the one with the bowed head. So the answer would be 3. Decide: Enough information is available to determine the correct option. Act: Output \texttt{<search>} to find the specific frame where the person is smiling and bowed. \texttt{<search>} person with head slightly bowed and smiling \texttt{</search>} \\

& \textbf{Video sample:}\\
&\includegraphics[width=0.8\linewidth]{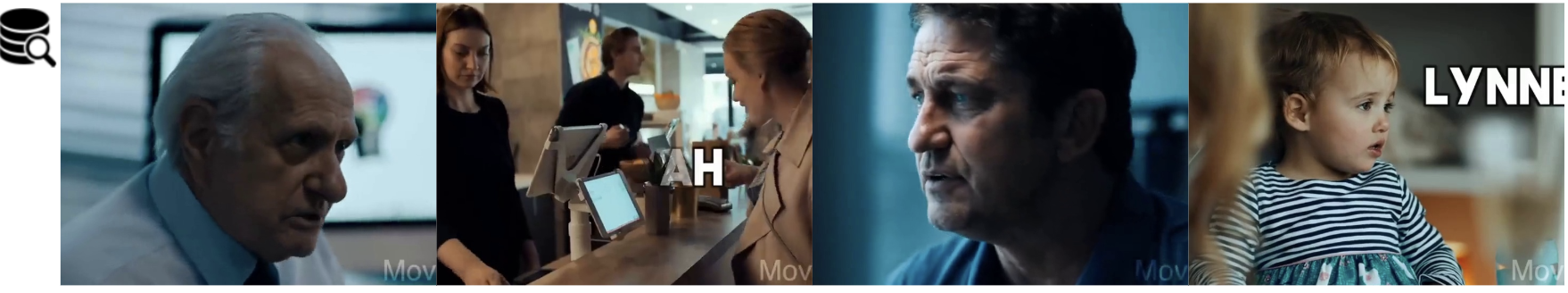}

\\
Turn 2 &
\textbf{Reasoning/output:} \texttt{<think>} From the frames, the woman in the scene with the head bowed and smiling is likely the one in the black dress (option 4), but also looking at the frames, there's a frame where a woman is at the counter, which aligns with the woman in the black dress. Alternatively, perhaps the woman in the olive-green coat (option 3) is the one. Wait, but the frames show a woman at the counter, which could be option 4. \texttt{</think>} Act: Confirm the correct option. \texttt{<search>} woman in the olive green coat\texttt{</search>} \\

&\textbf{Video sample:}\\
&\includegraphics[width=0.8\linewidth]{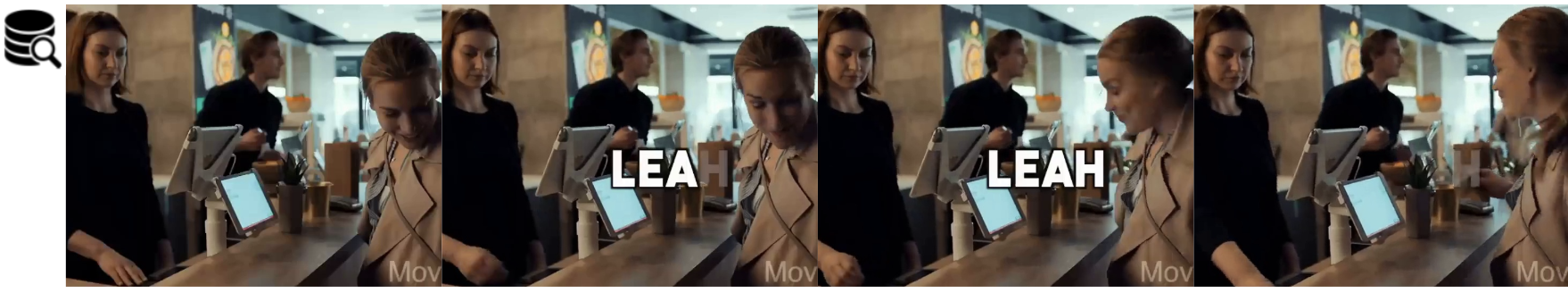}

\\
Turn 3 &
\textbf{Video sample:}~

\textbf{Reasoning/output:}
\texttt{<think>} Now, the frames show a woman with the black dress at the counter (option 4) but she is not smiling. The question matches the description of the woman in the olive green coat. So, the correct answer is 3.  \texttt{</think>} \texttt{<answer>} 3 \texttt{</answer>} \\

\textbf{Final Answer} & Option 3\\
\textbf{Ground Truth} & Option 3\\
\end{longtable}
\endgroup

\begingroup
\scriptsize
\setlength{\tabcolsep}{4pt}
\renewcommand{\arraystretch}{1.1}
\begin{longtable}{p{0.20\linewidth} p{0.77\linewidth}}
\caption{\texttt{VSeek} searching correctly to answer an event-specific grounding question}\label{tab:qual-trace-example5}\\
\toprule
\textbf{Turn / Field} & \textbf{Content} \\
\midrule
\endfirsthead

\toprule
\textbf{Turn / Field} & \textbf{Content} \\
\midrule
\endhead

\midrule
\multicolumn{2}{r}{\emph{Continued on next page}}\\
\endfoot

\bottomrule
\endlastfoot

Question & In a scene with white text 'I want you to start with that man over there, okay?' there are many people, and in front of a man in a black suit there is a triangular shelf with a basketball on it. What other objects are present in the scene?
\par\smallskip
\noindent 0. A doll wearing purple clothes
\par\smallskip
\noindent 1. A green plant
\par\smallskip
\noindent 2. A yellow incense burner
\par\smallskip
\noindent 3. A white chrysanthemum
\par\smallskip
\noindent 4. A white dress
\\
Original Video &
\includegraphics[width=1.0\linewidth]{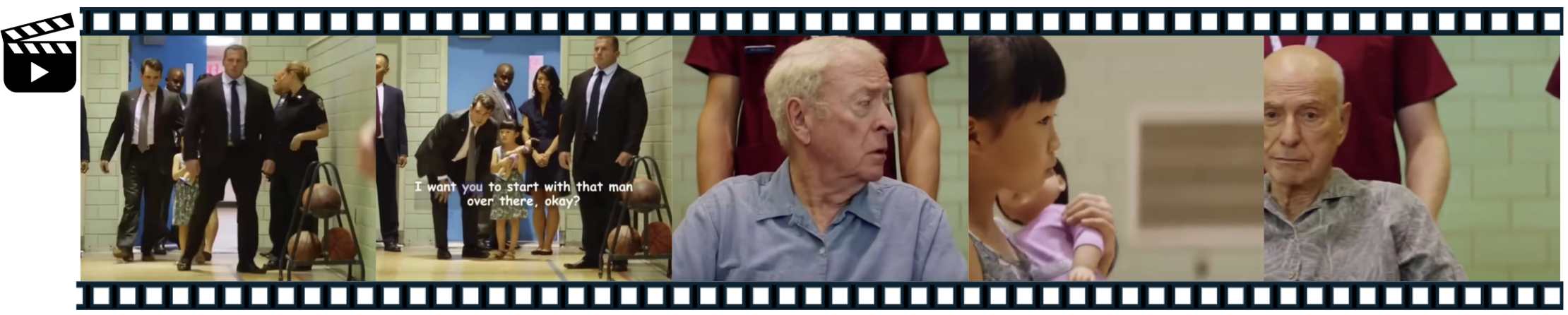}

\\
Turn 1 &
\textbf{Reasoning/output:} \texttt{<think>} Okay, let's try to figure out this problem. The user is asking which object is present in the scene where there's white text and a man in a black suit with a triangular shelf and a basketball. First, the options are 0 (doll in purple), 1 (green plant), 2 (yellow incense burner), 3 (white chrysanthemum), 4 (white dress). \\

& So, I need to check which of these is present. The key is to determine which object from the options is in the frames. Since the current frames might not show all, perhaps using video\_search with a language-based query could help. So I should call video\_search with a query that targets the possible objects. Maybe a search for the specific items in the scene. \\

& Alternatively, maybe the subtitle is already given, but the user mentioned the white text, so perhaps using a language-based search. So using video\_search with a query that includes the possible objects. Let me try searching for the objects present in the scene. Let me pick one of the options, but maybe the correct answer is one of them. Alternatively, perhaps the correct answer is one of the options, so using video\_search to find which one is there. \texttt{</think>} \texttt{<search>} a man in a black suit with a triangular shelf and a basketball \texttt{</search>} \\

& \textbf{Video sample:}\\
&\includegraphics[width=0.8\linewidth]{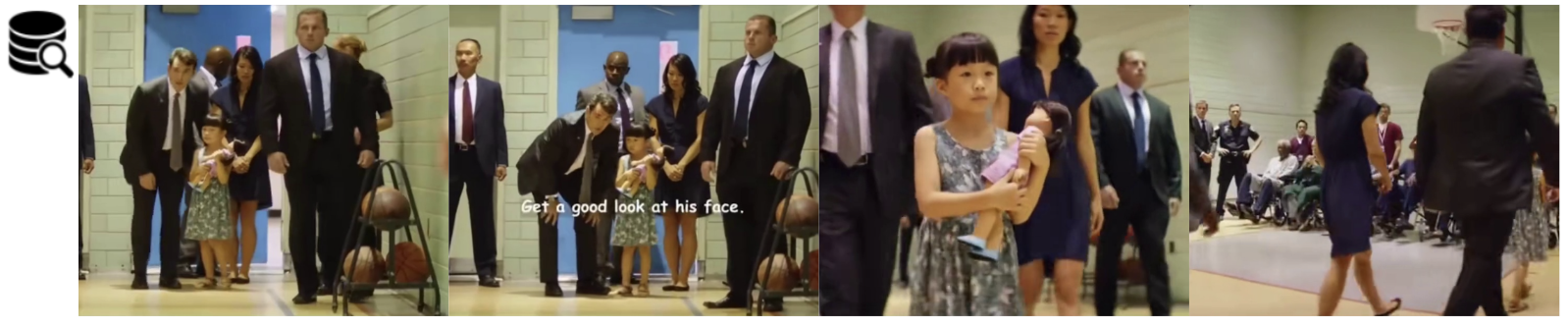}

\\
Turn 2 &
\textbf{Reasoning/output:} \texttt{<think>} Looking at the frames, there's a girl holding a pink object, which might be a doll. So option 0 is a doll wearing purple clothes. Alternatively, maybe the doll is present. Let me check again. The girl is holding a pink item, which could be a doll. So option 0 is correct. \texttt{</think>} \texttt{<search>} a doll wearing purple clothes \texttt{</search>} \\

&\textbf{Video sample:}\\
&\includegraphics[width=0.8\linewidth]{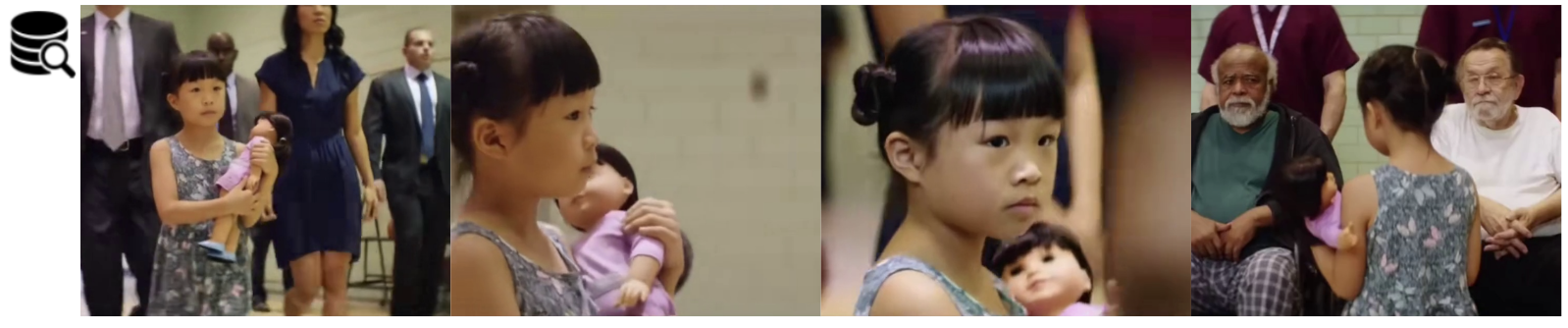}

\\
Turn 3 &
\textbf{Video sample:}~

\textbf{Reasoning/output:}
\texttt{<think>} Now it's clear that the girl is holding a doll, which matches option 0. So the correct answer is 0. \texttt{</think>} \texttt{<answer>} 0 \texttt{</answer>} \\

\textbf{Final Answer} & Option 0\\
\textbf{Ground Truth} & Option 0 \\
\end{longtable}
\endgroup

\subsection{Failure Cases}

\begingroup
\scriptsize
\setlength{\tabcolsep}{4pt}
\renewcommand{\arraystretch}{1.1}
\begin{longtable}{p{0.20\linewidth} p{0.77\linewidth}}
\caption{Qualitative example of a multi-turn reasoning trace}\label{tab:qual-trace-examplef1}\\
\toprule
\textbf{Turn / Field} & \textbf{Content} \\
\midrule
\endfirsthead

\toprule
\textbf{Turn / Field} & \textbf{Content} \\
\midrule
\endhead

\midrule
\multicolumn{2}{r}{\emph{Continued on next page}}\\
\endfoot

\bottomrule
\endlastfoot

Question & What type of movie is the scene in the video?
\par\smallskip
\noindent 0. Science Fiction
\par\smallskip
\noindent 1. Comedy
\par\smallskip
\noindent 2. Horror
\par\smallskip
\noindent 3. Documentary

\\
Original Video &
\includegraphics[width=1.0\linewidth]{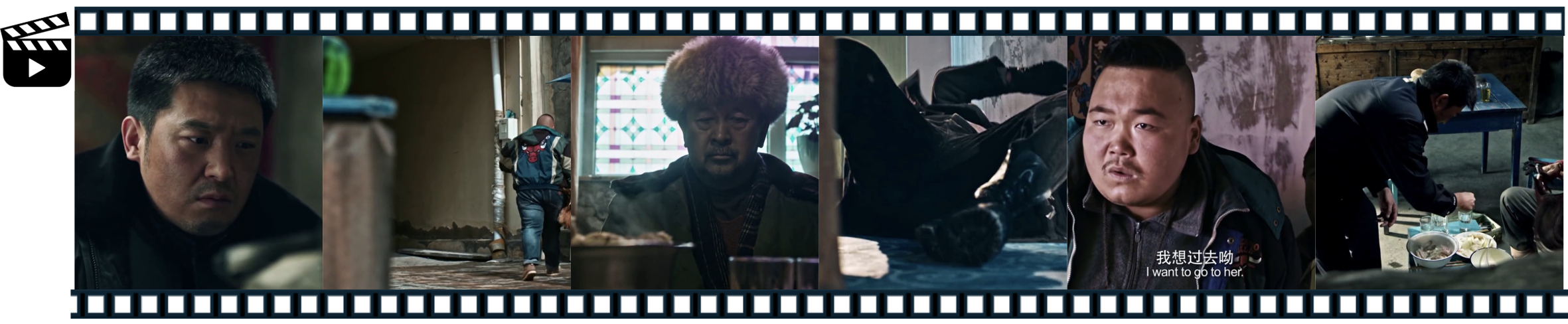}

\\
Turn 1 &
\textbf{Reasoning/output:} \texttt{<think>} Okay, let's try to figure out this problem. The user is asking for the type of movie the scene is in, with options 0 (Science Fiction), 1 (Comedy), 2 (Horror), 3 (Documentary). First, I need to determine which category the current video frames fall into. Since I don't have the actual video frames, I should use a tool to search for the relevant frames. The available tool is video\_search, which can take a query and a mode (base or subtitle). \\

& So, I should call video\_search with a query that helps identify the movie type. Maybe a language-based search, so mode is base. Let me use a query like "movie type" or "scene type". Alternatively, perhaps a specific action or object that's characteristic of the genre. Let me try using video\_search with the query "movie genre" and mode as base. That way, the tool will find the frames where the genre is evident, and then I can see which option matches. \texttt{</think>} \texttt{<search>} movie genre \texttt{</search>} \\

& \textbf{Video sample:}\\
&\includegraphics[width=0.8\linewidth]{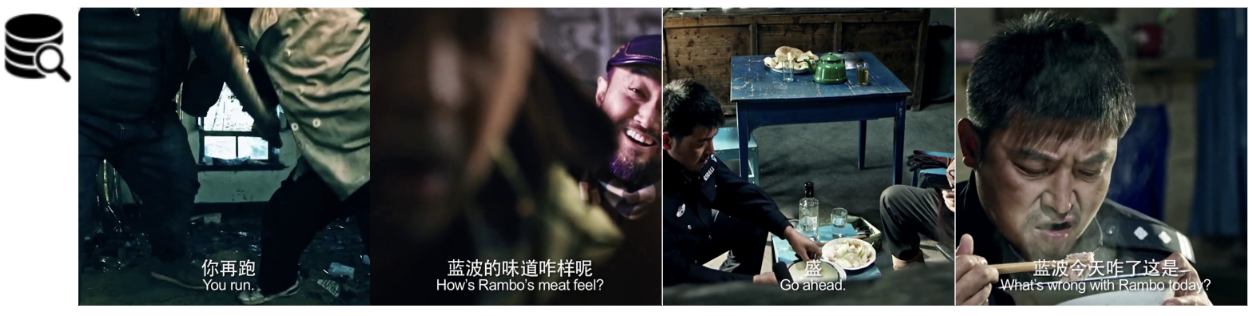}

\\
Turn 2 &
\textbf{Reasoning/output:} \texttt{<think>} Looking at the frames, there are subtitles in Chinese and English, and the setting seems to be a tense or dramatic conversation. The presence of characters in a dimly lit environment and the dialogue suggests it's a narrative-driven film. However, the options are 0 (Sci-Fi), 1 (Comedy), 2 (Horror), 3 (Documentary). Let me check which fits. \\

& Alternatively, maybe it's a documentary, but the subtitles and the style might not fit. Alternatively, perhaps it's a drama, but that's not an option here. Wait, the options are only 0,1,2,3. Alternatively, maybe it's a documentary (3), but the frames show people talking, which could be a documentary. Alternatively, perhaps it's a drama which might not be in the options. Hmm, maybe the correct answer is 3 (Documentary) or perhaps another. Alternatively, perhaps it's a thriller, which isn't an option, so maybe 3. \texttt{</think>} \texttt{<search>} documentary \texttt{</search>} \\

&\textbf{Video sample:}\\
&\includegraphics[width=0.8\linewidth]{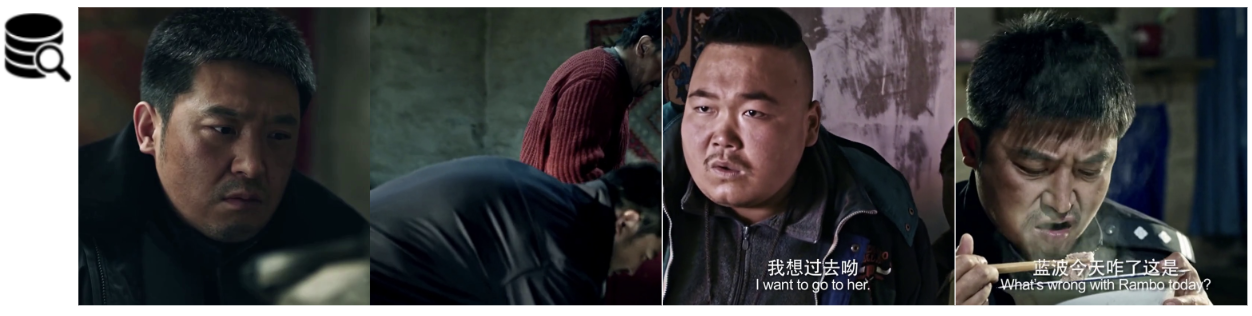}

\\
Turn 3 &

\textbf{Reasoning/output:}
\texttt{<think>} Now, the frames show a scene with people and a dog, and the subtitles are in Chinese and English. The presence of a dog and the dialogue might indicate a documentary or perhaps a realistic film, but the options are 0,1,2,3. Alternatively, maybe it's a documentary (3). Alternatively, perhaps it's a drama, but the closest option is 3. Let me check again. Alternatively, perhaps it's a Comedy (1), but the tone seems more serious. So the answer is 3. \texttt{</think>} \texttt{<answer>} 3 \texttt{</answer>} \\

\textbf{Final Answer} & Option 1\\
\textbf{Ground Truth} & Option 3 \\
\end{longtable}
\endgroup

\begingroup
\scriptsize
\setlength{\tabcolsep}{4pt}
\renewcommand{\arraystretch}{1.1}
\begin{longtable}{p{0.20\linewidth} p{0.77\linewidth}}
\caption{Answering incorrectly without searching}\label{tab:qual-trace-example7}\\
\toprule
\textbf{Turn / Field} & \textbf{Content} \\
\midrule
\endfirsthead

\toprule
\textbf{Turn / Field} & \textbf{Content} \\
\midrule
\endhead

\midrule
\multicolumn{2}{r}{\emph{Continued on next page}}\\
\endfoot

\bottomrule
\endlastfoot

Question & Next to the swimming pool, there are white sunshades and trees, and there are some people inside the pool. A man wearing a gray short-sleeve shirt is taking pictures in front of the outdoor swimming pool. Besides the swimming pool, there is also a wooden dining table and some chairs. Which scene appears first?
\par\smallskip
\noindent 0. Next to the swimming pool, there are white sunshades and trees, and there are some people inside the pool, appears first; A man wearing a blue short-sleeve shirt is taking pictures in front of the outdoor swimming pool. Besides the swimming pool, there is also a wooden dining table and some chairs, appears later.
\par\smallskip
\noindent 1. They are one scene.
\par\smallskip
\noindent 2. Next to the swimming pool, there are white sunshades and trees, and there are some people inside the pool, appears first; A man wearing a gray short-sleeve shirt is taking pictures in front of the outdoor swimming pool. Besides the swimming pool, there is also a wooden dining table and some chairs, appears later.
\par\smallskip
\noindent 3. A man wearing a gray short-sleeve shirt is taking pictures in front of the outdoor swimming pool. Besides the swimming pool, there is also a wooden dining table and some chairs, appears first. Next to the swimming pool, there are white sunshades and trees, and there are some people inside the pool, appears later.
\\
Original Video &
\includegraphics[width=1.0\linewidth]{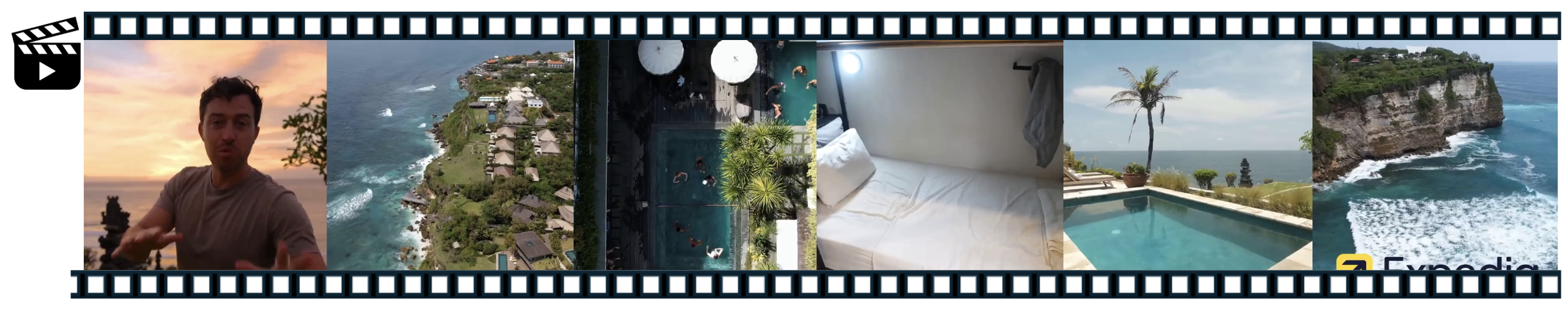}

\\
Turn 1 &
\textbf{Reasoning/output:} \texttt{<think>} Okay, let's try to figure out this problem. The question is asking which scene appears first between the two parts described in the options. First, let me read the options again. Option 0 says: The part where there are white sunshades, trees, people in the pool appears first, and then the man in the gray shirt taking pictures, plus the dining table and chairs. \\

& Option 2 says the man taking pictures and the dining table come first, then the sunshades and people. But according to the user's description, the correct scenario is that the white sunshades and trees with people in the pool are first, then the man taking pictures and the dining table. Looking at the options, Option 2 is the one where the man and the dining table come first, which is not correct. Option 0 has the first part as the sunshades and people, then the man, which matches the description. \\

& So the correct answer should be 0? Wait, but the options are 0, 1, 2, 3. Wait the user's options are: 0) Next to the pool, sunshades and trees, people inside pool (first); then man in blue shirt taking pictures, etc. (but the man in the problem is gray, so maybe this is a typo?) Wait the problem states the man is wearing a gray short-sleeve shirt. So in option 0, the man is in a blue shirt, which is different. So perhaps option 0 is incorrect. \\

& Option 2: Next to the pool, sunshades and trees, people (first); then the man in gray shirt and dining table. So that matches the problem's description. Wait, the problem says: "Next to the swimming pool, there are white sunshades and trees, and there are some people inside the pool. A man wearing a gray short-sleeve shirt is taking pictures in front of the outdoor swimming pool. Besides the swimming pool, there is also a wooden dining table and some chairs." \\

& So the first part (sunshades, trees, people) comes first, then the man and the dining table. So the correct option is 2. Wait, looking at the options: Option 2: "Next to the swimming pool, there are white sunshades and trees, and there are some people inside the pool, appears first; A man wearing a gray short-sleeve shirt is taking pictures in front of the outdoor swimming pool. Besides the swimming pool, there is also a wooden dining table and some chairs, appears later." \\

& Yes, that's exactly the case. So the correct answer is 2. \texttt{</think>} Reason: The correct scenario is that the sunshades, trees, and people in the pool appear first, followed by the man taking pictures and the dining table. This matches option 2. Decide: Enough information is available to confirm the correct option. Act: Output the correct option number. \texttt{</think>} \texttt{<answer>} 2 \texttt{</answer>} \\

\textbf{Final Answer} & Option 2\\
\textbf{Ground Truth} & Option 0 \\
\end{longtable}
\endgroup

\end{document}